\documentclass{article} 
\usepackage{iclr2025_conference,times}

\usepackage{hyperref}
\usepackage{url}

\def\eqref#1{equation~\ref{#1}}

\def\1{\bm{1}}

\DeclareMathAlphabet{\mathsfit}{\encodingdefault}{\sfdefault}{m}{sl}
\SetMathAlphabet{\mathsfit}{bold}{\encodingdefault}{\sfdefault}{bx}{n}

\usepackage{hyperref}
\usepackage{url}
\usepackage{graphicx}
\graphicspath{images/}
\usepackage{blindtext}
\usepackage{amsmath,amsfonts}
\usepackage{algorithmic}
\usepackage{algorithm}
\usepackage{array}
\usepackage[caption=false,font=normalsize,labelfont=sf,textfont=sf]{subfig}
\usepackage{textcomp}
\usepackage{stfloats}
\usepackage{url}
\usepackage{verbatim}
\usepackage{cite}
\usepackage{lipsum}
\usepackage{bbm}
\usepackage{makecell}
\usepackage{stfloats}
\usepackage{enumitem}
\usepackage{tablefootnote}
\usepackage{float}
\usepackage{amsmath,bm}
\usepackage{xcolor}
\usepackage{stmaryrd}
\usepackage{setspace}
\usepackage{alphalph,etoolbox}
\usepackage{orcidlink}
\usepackage{booktabs}
\usepackage{todonotes}
\usepackage{xcolor}
\usepackage{CJKutf8}
\usepackage{hyperref}

\newif\ifhighlight
\highlightfalse
\ifhighlight

\newcommand{\yw}[1]{\textcolor{purple}{#1}}
\newcommand{\ywz}[1]{\textcolor{red}{#1}}

\else 

\newcommand{\yw}[1]{{#1}}
\newcommand{\ywz}[1]{{#1}}

\fi

\title{CAPM: Fast and Robust Verification  on Maxpool-based CNN via  Dual Network}

\author{Jia-Hau Bai\\ 
Graduate Institute of Communication Engineering, National Taiwan University, Taipei, Taiwan \\
\texttt{r07942092@ntu.edu.tw
}
\AND
Chi-Ting Liu\\ 
Department of Electrical Engineering, National Taiwan University, Taipei, Taiwan \\
\texttt{b07901090@ntu.edu.tw}
\AND
Yu Wang\\
Graduate Institute of Communication Engineering, National Taiwan University, Taipei, Taiwan \\
\texttt{r11942152@g.ntu.edu.tw}
\AND
Fu-Chieh Chang \\
MediaTek Research, Taipei, Taiwan \\
Graduate Institute of Communication Engineering, National Taiwan University, Taipei, Taiwan \\
\texttt{d09942015@ntu.edu.tw}
\AND
Pei-Yuan Wu  \\
Graduate Institute of Communication Engineering, National Taiwan University, Taipei, Taiwan \\
\texttt{peiyuanwu@ntu.edu.tw}
}

\iclrfinalcopy 
\begin{document}

\maketitle

\begin{abstract}
This study uses CAPM \yw{(Convex Adversarial Polytope for Maxpool-based
CNN)} to improve the verified bound for general purpose maxpool-based \yw{convolutional neural networks (CNNs)} under bounded norm adversarial perturbations. The maxpool function is decomposed as a series of ReLU functions to extend the convex relaxation technique to maxpool functions,  by which the verified bound can be efficiently computed through a dual network.
The experimental results demonstrate that this technique allows the state-of-the-art verification precision for maxpool-based CNNs and involves a much lower computational cost than current verification methods, such as DeepZ, DeepPoly and PRIMA. This method is also applicable to large-scale CNNs, which previous studies show to be often computationally prohibitively expensive. 
Under certain circumstances, CAPM is 40-times, 20-times or twice as fast and give a significantly higher verification bound (CAPM 98\% vs. PRIMA 76\%/DeepPoly 73\%/DeepZ 8\%) as compared to PRIMA/DeepPoly/DeepZ. (cf. Fig.~\ref{fig:compare cifar10} and Fig.~\ref{fig:time cifar10}).
\yw{Furthermore, we additionally present the time complexity of our algorithm as $O(W^2NK)$, where $W$ is the maximum width of the neural network, $N$ is the number of neurons, and $K$ is the size of the maxpool layer's kernel.}
\end{abstract}

\section{Introduction}

In the past few years, convolution neural networks have reached unprecedented performance in various tasks such as face recognition \citep{hu2015face,mehdipour2016comprehensive} and self-driving cars \citep{rao2018deep,maqueda2018event}, to name a few.
However, these networks are vulnerable to malicious modification of the pixels in input images, known as adversarial examples, such as FGSM \citep{goodfellow2014explaining}, PGD \citep{madry2017towards}, One Pixel Attack \citep{su2019one}, Deepfool \citep{moosavi2016deepfool}, EAD \citep{chen2018ead}, GAP \citep{poursaeed2018generative}, MaF\citep{chaturvedi2020mimic} and many others \citep{wong2019wasserstein}.

In view of the threat posed by adversarial examples, how
to protect neural networks from being tricked by adversarial
examples has become an emerging research topic. Previous studies of defense against adversarial examples are categorized as \textit{Removal of adversarial perturbation} \citep{akhtar2018defense, xie2019feature, jia2019comdefend, samangouei2018defense} and \textit{Adversarial training} \citep{shafahi2019adversarial, han2020way, tramer2020adaptive}. Both defense mechanisms may protect the network from certain adversarial examples but there is only empirical evidence that they do so. The robustness of the network is not guaranteed. It is impossible to train or evaluate all possible adversarial examples so these methods are vulnerable to other adversarial examples that are not in the data sets that are used.

The need for guaranteed robustness assessments has led to the development of  verification mechanisms for a neural network. These verify specific properties pertaining to neural networks, such as robustness against norm-bounded perturbation \citep{dvijotham2018dual, singh2018fast}, robustness against adversarial frequency or severity \citep{katz2017reluplex} and robustness against rotations \citep{singh2019abstract}.
%

During the early development of neural network verification, satisfiability modulo theories (SMT) solver \citep{katz2017reluplex} and semidefinite programming (SDP) methods \citep{raghunathan2018certified} were used. A SMT solver yields tight verification bounds but is not scalable to contemporary networks with sophisticated architecture. The SDP method requires less time but is limited to linear architectures. Recent studies have developed verification tools for more realistic scenarios, such as a fully connected neural network (FCNN) with an activation function and a convolution neural network (CNN). As indicated by \citet{salman2020convex}, the main methods for neural network verification can be categorized as either primal view or dual view.
%

The primal view method involves \textit{Abstract interpretation} and \textit{Interval bound propagation}. There are classic frameworks for abstract interpretation (e.g., AI2 \citep{gehr2018ai2}, DeepZ \citep{singh2018fast}, DeepPoly \citep{singh2019abstract}). As a step further,  RefineZono \citep{singh2019boosting} and RefinePoly \citep{singh2019boosting} use mixed integer linear programming (MILP) to improve the verification bounds for DeepZ and DeepPoly, respectively. However, the computation time that is required for verification is significantly increased. Another bounding technique for the primal view method is interval bound propagation, which uses interval arithmetic to obtain the bound for each individual neuron in each layer. Representative works include IBP \citep{gowal2018effectiveness} and CROWN-IBP \citep{zhang2019towards}.  
Dual-view methods \citep{wong2018provable,dvijotham2018dual,wong2018scaling,bunel2020lagrangian,DBLP:journals/corr/abs-2011-13824,NEURIPS2021_fac7fead} formulate the verification problem as an optimization problem, so according to Lagrangian duality \citep{boyd2004convex}, each dual feasible solution yields a lower bound to the primal problem and verification bounds are derived by solving the dual problem. \yw{Moreover, noteworthy among these methods is the state-of-the-art approach $\alpha$,$\beta$-CROWN \citep{NEURIPS2021_fac7fead}, which is grounded in the LiRPA framework \citep{DBLP:journals/corr/abs-2011-13824}.}
%

\subsection{Verification of a CNN {with} maxpooling}\label{subsec:Verification of CNN with maxpooling}

The maxpool function is an integral part in most real-world neural network architectures, especially CNNs (e.g., LeNet \citep{lecun1998gradient}, AlexNet \citep{krizhevsky2012imagenet} and VGG \citep{simonyan2014very}), which are widely used for image classification. However, past works have the following shortage in the verification of networks involving maxpool functions:

\begin{itemize}
    \item \textit{Not applicable:} 
    IBP \citep{gowal2018effectiveness} and others \citep{wong2018provable,wong2018scaling} \citep{bunel2020lagrangian,de2021improved} verify a CNN but there is no theory to verify maxpool-based networks. \citet{gowal2018effectiveness} analyzed several monotonic activation functions (e.g., ReLU, tanh, sigmoid) in IBP but they did not consider non-monotonic functions (e.g., maxpool). \citet{wong2018provable} discussed the verification of a ReLU-based FCNN and a later study \citep{wong2018scaling} uses this for the verification of residual networks. However, besides referring to the work of \citet{dvijotham2018dual}, the study by \citet{wong2018scaling} does not address much about handling maxpool functions. \citet{bunel2020lagrangian,de2021improved} analyzed networks with nonlinear activation functions, such as ReLU and sigmoid, but there is no analysis of the maxpool function.
    %
    %
    \item \textit{Has theory but lack of implementation evidence:}
    These studies analyze the maxpool function but experiments only verifiy ReLU-based CNNs. Examples include AI2 \citep{gehr2018ai2}, DeepZ \citep{singh2018fast}, DeepPoly \citep{singh2019abstract}, RefineZono \citep{singh2019boosting}, RefinePoly \citep{singh2019boosting} and LiRPA \citep{DBLP:journals/corr/abs-2011-13824}. \citet{dvijotham2018dual} analyzed a large variety of activation functions, such as ReLU, tanh, sigmoid and maxpool, {but they only demonstrated the experiment result on} a small network consisting of one linear layer, followed by sigmoid and tanh. CROWN-IBP \citep{zhang2019towards} used the verification method of IBP \citep{gowal2018effectiveness} and analyzed non-monotonic functions, including maxpool, but experiments only verify results for ReLU-based CNNs. \yw{In spite of providing functions corresponding to maxpool, LIRPA is currently unable to function properly on maxpool-based CNNs. The definitions of DenseNet and ResNeXt used in their experiments can be found in their GitHub repository, and it's noteworthy that these definitions do not include a maxpool layer.}
    %
    %
    %
    \item \textit{Imprecise:}
    These studies give an imprecise verification bound for maxpool-based CNN. DeepZ \citep{singh2018fast}, DeepPoly \citep{singh2019abstract}, and PRIMA \citep{muller2022prima} implement the verification of maxpool-based CNN but experimental results are nevertheless lacking. For comparison purposes, we implement these methods on 6 maxpool-based CNN benchmarks modified from \citet{mirman2018differentiable} (cf. Supplementary Material \ref{appendices: network structure}). Our experiment indicates that DeepZ, DeepPoly, and PRIMA are imprecise in our benchmarks. For a norm-bounded perturbation $\epsilon=0.0024$, the verified robustness for a convSmall CIFAR10 structure decreases to 1\%, 25\%, and 26\%, respectively (cf. Fig.~\ref{fig:compare cifar10}). \yw{Due to the implementation of the BaB (branch and bound) algorithm in $\alpha$,$\beta$-CROWN \citep{NEURIPS2021_fac7fead} on maxpool-based CNNs, it results in excessive GPU memory requirements (more than 13GB) or prolonged execution times (more than 5 minutes per example). Consequently, we consider examples that trigger the BaB algorithm as not verified.}
    %
    %
    \item \textit{Computational costly:} 
    These studies involve a significant computational cost to verify each input image in the maxpool-based CNN benchmarks (cf. Sec.~\ref{sec:experiment_main}). Our experiment shows that for $\epsilon = 0.0006$ on convBig CIFAR10 (cf. Fig.~\ref{fig:time cifar10}), PRIMA \citep{muller2022prima} requires 6.5 days and DeepPoly \citep{singh2019abstract} requires 3 days to verify 100 images. (See Sec.~\ref{subsec:experiment setting} for hardware spec)
    %
\end{itemize}

\citet{yuan2019adversarial} showed that $l_\infty$ norm is one of the most commonly used perturbation measurements (e.g.,DeepFool \citep{moosavi2016deepfool}, CW \citep{carlini2017towards}, Universal adversarial perturbations \citep{moosavi2017universal}, and MI-FGSM \citep{dong2018boosting}). Therefore, this study verifies a maxpool-based CNN with $l_\infty$ norm-bounded perturbation. The contributions of this study are:

\begin{itemize}
    \item 
    CAPM (Convex Adversarial Polytope for Maxpool-based CNN) is used to improve the verified bound for a maxpool-based CNN, assuming $l_\infty$ norm-bounded input perturbations. The maxpool function is decomposed into multiple ReLU functions (see Sec.~\ref{subsection:Overview of our method}), as such convex relaxation trick by \citet{wong2018provable} can be applied. \yw{While we are not the first to consider the relationship between maxpool functions and ReLU functions in our paper, our experiments demonstrate that the algorithm we propose is currently the most effective among all executable methods.}
    %
    %
    \item
    A dual network for general purpose CNNs is derived {with} maxpool, padding and striding operations to {obtain} the verified bound efficiently (cf. \eqref{dual network}).
    %
    %
    \item
    The results (cf. Sec.~\ref{sec:experiment_main}) show that CAPM gives a verification that is as robust as the state-of-the-art methods (PRIMA), but there is significantly less runtime cost for MNIST and CIFAR 10.
    %
    %
    \item 
    The experimental results show the limitations of a Monte-Carlo simulation for a large-scale neural network verification problem, which demonstrates the necessity for a provable robustness verification scheme.
    %
    \item 
    \yw{Among algorithms applicable to maxpool-based CNNs, we are currently the only ones providing a clearly defined time complexity for our algorithm. The time complexity of the CAPM algorithm is $O(W^2NK)$.}
\end{itemize}

\yw{The main limitation of our algorithm is that it requires adherence to the assumptions defined in Supplementary Material.~\ref{subsubsec:Notation and data flow}. We address future work based on this in the conclusions section.}

The reminder of this paper is organized as follows: Sec.~\ref{section:method} defines the verification problem and introduce the key idea of CAPM with a toy example. Sec.~\ref{sec:experiment_main} compares the experimental result for CAPM  \yw{against with DeepZ, DeepPoly, PRIMA, LiRPA and $\alpha$,$\beta$-CROWN} in terms of the verified robustness and average runtime metrics for various adversary budgets. The experiment result also indicates that the accuracy lower bound predicted by each verification method becomes looser as norm-bounded perturbation increases. Conclusions and future work are summarized in Sec.~\ref{section:conclusion}.
Furthermore, we gives a bound analysis for intermediate layers to explain the reason for this phenomenon in Supplementary Material  \ref{section:Bound Analysis for Intermediate Layers}. 
\section{Method}
\label{section:method}

This section illustrates the verification of a CNN classifier under the $l_\infty$ norm-bounded perturbations. Sec.~\ref{subsec:Verification Problem} defines the verification problem and Sec.~\ref{subsection:Overview of our method} gives an overview of CAPM. More details for solving the verification problem for a maxpool-based CNN are in Supplementary Material \ref{subsec:Maxpool-based CNN Architecture} and Supplementary Material \ref{subsec:Primal problem}.
%

Supplementary Material \ref{subsec:Maxpool-based CNN Architecture} specifies the CNN architecture for this study; Supplementary Material \ref{subsec:Primal problem} describes the formulation of the the verification problem in terms of Lagranian dual problems, and simplifies the dual constraints to the form of a leaky ReLU dual network.
%

\subsection{Definition of the verification problem}\label{subsec:Verification Problem}

\begin{figure}[ht]
\centering
\includegraphics[width=0.8\textwidth]{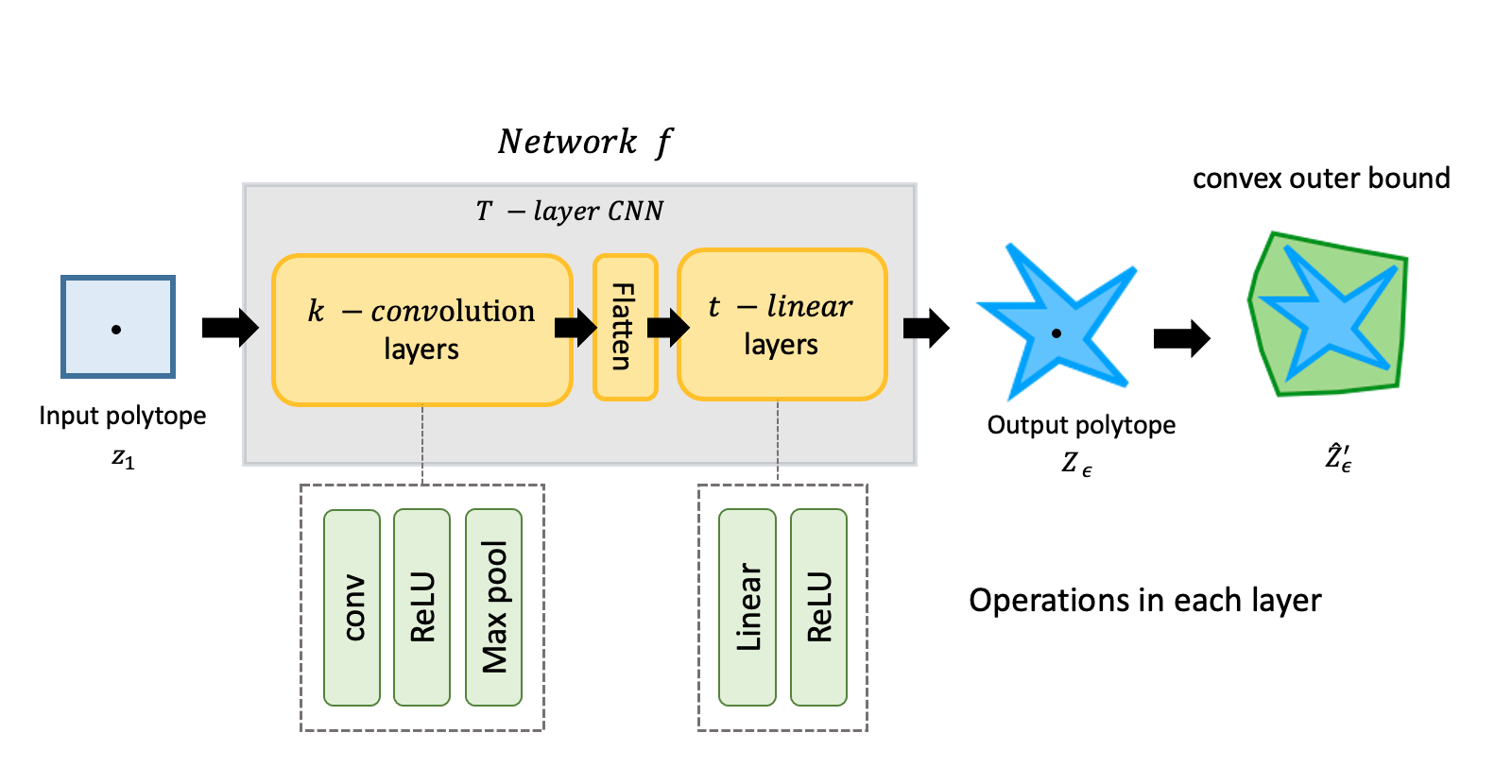}
\caption{\label{fig:3.1_output_polytope.png} Adversarial polytope passes through network $f$}
\end{figure}

This study evaluates the robustness of a CNN to arbitrary adversarial examples within a bounded norm budget, to see whether or not the prediction result of the CNN will change under such adversarial perturbations. The verification problem is reformulated through convex relaxation as a convex optimization problem and the duality theorem states that any dual feasible solution can serve as a lower bound for the original verification problem.
%

A neural network $f$ consisting of $k$ convolution layers with ReLU activation and maxpool functions, followed by flattening and $t$ fully connected layers with ReLU activation is shown in Fig
\ref{fig:3.1_output_polytope.png}.	If a clean image $x$ is added with perturbation $\Delta$, to which this study impose a $l_\infty$-norm
constraint $\lVert \Delta \rVert_\infty  \leq \epsilon$, the perturbed input $z_1$ resides in an input adversarial polytope that is described as:
%
\begin{equation}\label{noise_def}
	\bf{z_1} \leq \bf{x} + \bf{\epsilon} \quad \text{ and }\quad
	\bf{z_1}  \geq \bf{x} - \bf{\epsilon} .
\end{equation}
The perturbed input $z_1$ is taken as the input of the network $f$. \ywz{Though at first glimpse the input adversarial polytope (cf. \eqref{noise_def}) is a hyper-cube which is convex, and that the ReLU function is itself a convex function, the intermediate adversarial polytope after passing the input adversarial polytope through the convolution and ReLU activation is in general not a convex set, as the set $\{(\xi,ReLU(\xi)): \xi \in \mathbb{R}\}$ is not a convex subset of $\mathbb{R}^2$.}
\yw{The verification problem deviates from a convex optimization framework due to the non-linearity of ReLU and maxpool equalities. The presence of these non-linear equalities introduces complexity, rendering the optimization problem non-convex. Similarly, the application of a nonlinear function leads to a non-convex output polytope, even if the input polytope maintains convexity. In other words, substituting nonlinear equality constraints with linear inequality constraints results in the creation of a convex outer bound.}
To make the feasible solution a convex set, a convex outer bound \citep{wong2018provable} is constructed 
when passing through each of the ReLU and maxpool functions. The output polytope $Z_\epsilon$, 
which is the collection of all possible results computed by $f$ at the
output layer with a perturbed input $z_1$, is contained within a convex polytope $\hat{Z}'_\epsilon$ . Both $Z_\epsilon$
and $\hat{Z}'_\epsilon$ are subsets of $\mathbb{R}^K$ for a $K$-class classification task.
%

For an image $x$ that is labeled with ground truth $y^* \in \{1,...,K\}$ to which an adversary attempts to mislead network $f$ into falsely predicting a target label $y^{targ} \in \{1,...,K\}$ rather than $y^*$, a necessary condition is that the adversary must find a perturbed input $z_1$ satisfying \eqref{noise_def} so that ${\bf e}_{y^*}^T f(z_1) \leq {\bf e}_{y^{targ}}^T f(z_1)$, where ${\bf e}_{y^*}$ and ${\bf e}_{y^{targ}}$ are one-hot encoded vectors of $y^*$ and $y^{targ}$, respectively. Therefore, as $f(z_1) \in Z_\epsilon \subset \hat{Z}'_\epsilon$, if the minimum for the optimization problem
%
\begin{equation}\label{objective}
    \begin{aligned}
        \min_{\bf{\hat{y}} \in \bf{\hat{Z}'_\epsilon}} \quad & (\bf{e_{y*}}-\bf{e_{y^{targ}}})^T\bf{\hat{y}}
    \end{aligned}
\end{equation}
is positive for every target class $y^{targ} \in \{1,...,K\} \setminus \{y^*\}$, then network $f$ cannot be fooled by an adversarial example that differs from image $x$ by a perturbation with at most $\epsilon$ under $l_\infty$-norm.
This method can guarantee zero false negatives, so the system flags every image that is prone to attack by an adversarial example, but it may falsely flag some images resilient to perturbations.
%

\subsection{CAPM overview}
\label{subsection:Overview of our method}

This section describes a toy example to illustrate the use of CAPM to solve the optimization problem in \eqref{objective} using a Maxpool-based network. The dual problem is formulated using Lagrangian relaxation \citep{boyd2004convex} and convex relaxation \citep{wong2018provable}, so any dual feasible solution corresponds to a lower bound to the original problem in \eqref{objective}. Convex relaxation loosens the verification bound but \citet{wong2018provable} showed that this lower bound can be calculated using a \ywz{backpropagation-like} \yw{dynamic programming} process in a so called dual network. \yw{As the determination of upper and lower bounds for preceding layers constitutes a sub-problem within the dual network framework for subsequent layers, employing a dynamic programming algorithm becomes a viable approach to address this verification problem.} This study extends the method of \citet{wong2018provable} to a maxpool-based CNN and demonstrates that maxpool-based CNNs can also be verified efficiently and precisely using a dual network.
%
\subsubsection{Toy example}

CAPM verifies the robustness of a simple maxpool-based network under the $l_\infty$ norm constraint in \eqref{noise_def}. The verification problem for this toy example is formulated as an optimization problem
in \eqref{eq:toy_ex_problem}. If the lower bound of \eqref{eq:toy_ex_problem} is positive for all possible target classes, then this network is not misled by any input perturbation, $l_\infty$-norm that is less than $\epsilon$. If not, then this network may not be safe for this input perturbation.
%
\begin{equation}\label{eq:toy_ex_problem}
    \begin{aligned}
        \min_{ \hat{\mathbf{z}}'_3 } \quad & (\bf{e_{y*}}-\bf{e_{y^{targ}}})^T \hat{\mathbf{z}}_3 \equiv \mathbf{d}^T \hat{\mathbf{z}}_3\\
        \textrm{s.t.} \quad & \mathbf{z}_1 \leq \bf{x}+\bf{\epsilon}\\
                      \quad & \mathbf{z}_1 \geq \bf{x}-\bf{\epsilon}\\
                      \quad & \hat{\mathbf{z}}_2 = \mathbf{W}_1\mathbf{z}_1 + \mathbf{b}_1 \\
                      \quad & \mathbf{z}_2^R = \max (\hat{\mathbf{z}}_2, \  \mathbf{0}) \\
                      \quad & \mathbf{z}_2 = \max ( z^R_{2,0}, \ z^R_{2,1}, \  z^R_{2,2}, \ z^R_{2,3}) \\
                      \quad & \hat{\mathbf{z}}_3 = \mathbf{W}_2\mathbf{z}_2 + \mathbf{b}_2 \\
    \end{aligned}
\end{equation}
In \eqref{eq:toy_ex_problem}, the perturbed input $\mathbf{z}_1$ is the input to a simple maxpool-based network. The feature map $\hat{\mathbf{z}}_2$ is obtained by inputting $\mathbf{z}_1$ into the linear operation in the fully-connected layer. ReLU and maxpool are then used to compute the intermediate results $\mathbf{z}^R_2$ and $\mathbf{z}_2$. The output $\hat{\mathbf{z}}_3$ is calculated using the linear operation. 
This is a non-convex optimization problem because of the non-affine activation functions ReLU and maxpool so \citet{wong2018provable}'s method of convex relaxation (see Supplementary Material~\ref{subsubsec:Convex Outer Bound} for more details) is applied to the ReLU function over the input interval, which approximates the ReLU function using the linear outer bounds (cf. Fig.~\ref{fig:convex outer bound.png}). In terms of the maxpool function (see \ref{subsubsec:Notation and data flow} and \ref{subsubsec:Convex Outer Bound} for more details), $z_2 = \max ( z^R_{2,0}, \ z^R_{2,1}, \  z^R_{2,2}, \ z^R_{2,3})$ is decomposed into several one-by-one comparisons using dummy variables
%
\begin{equation}\label{eq:toy_ex_maxpool relaxation}
    \begin{aligned}
        z^M_{2,j+1} &= \max (z^R_{2,j}, \ z^M_{2,j}) = z^M_{2,j} + \max(z^R_{2,j}-z^M_{2,j}, \ 0),\  j \in \llbracket 0, 3\rrbracket.
    \end{aligned}
\end{equation}
As such, $z^M_{2,4} = \max(z^M_{2,0}, z^R_{2,0}, z^R_{2,1}, z^R_{2,2}, z^R_{2,3})$, and $z_{2} = z^M_{2,4}$ if one chooses $z^M_{2,0}$ no larger than the maximum of $z^R_{2,0},...,z^R_{2,3}$. In this example, $z^M_{2,0} = 0$ because all elements in $z^R_2$ are the output of ReLU so they are non-negative, \eqref{eq:toy_ex_maxpool relaxation} is then split into several terms:
\begin{subequations}\label{eq:toy_ex_maxpool seperate}
    \begin{align}
        \begin{split}
            &\bar{z}_{2,j} = z^R_{2,j} - z^M_{2,j} 
        \end{split}\label{toy_ex_maxpool seperate - a}\\
        \begin{split}
            &z'_{2,j} = \max(\bar{z}_{2,j}, 0) 
        \end{split}\label{toy_ex_maxpool seperate - b}\\
        \begin{split}
            &z^M_{2,j+1} = z'_{2,j} + z^M_{2,j} 
        \end{split}\label{toy_ex_maxpool seperate - c}
    \end{align}
\end{subequations}
After decomposition using \eqref{eq:toy_ex_maxpool seperate}, the second term is also a ReLU function so convex relaxation is used, assuming knowledge of the upper-lower bounds of $\bar{z}_{2,j}$. The ReLU and maxpool functions in \eqref{eq:toy_ex_problem} are then replaced by \ywz{convex outer bounds} \yw{linear inequalitiy constraints} to form a convex optimization problem. {The dual problem can then be written in the form of a dual network (see Supplementary Material \ref{subsubsec:Dual Problem} for detailed derivation):} 
%
\begin{equation}\label{eq:toy_ex_dual network}
    \begin{aligned}
        \max J_\epsilon(\Theta) \quad
        \textrm{s.t.} \quad  F(\mathbf{d}, \mathbf{\alpha}^R, \mathbf{\alpha}^M)
    \end{aligned}
\end{equation}
\ywz{which has the form of a leaky-ReLU network}.
\yw{If the dual optimal of the convex relaxation problem is positive,} then the system verifies the network as robust and if not, the system does not exclude the possibility that the network can be fooled by some perturbation with $l_\infty$-norm of at most $\epsilon$. \ywz{The variables $\alpha^R$ and $\alpha^M$ are considered to be additional free variables for the dual network, the choice of which affects the precision of the verification bound that is calculated by the dual network. A strategy that is similar to CAP \citep{wong2018provable} is used to determine $\alpha^R$ and $\alpha^M$.}
%
\subsubsection{Determining the upper-lower bound}
\label{subsubsec:Determining the upper-lower bound}
The node-wise upper-lower bounds are required for the convex relaxation of ReLU functions. To determine the upper-lower bounds, namely $\hat{{l}}_{2,j} \leq {\hat{{z}}_{2,j}} \leq \hat{{u}}_{2,j}$, for the input nodes of the ReLU function, a verification problem is formulated that corresponds to the network up to the linear layer before the first ReLU. The node-wise bounds for ${\hat{{z}}_{2,j}}$ are determined by evaluating the resulting (smaller) dual network with one-hot input vector $\bm{e}_j$ (instead of $\mathbf{d}$) \citep{wong2018provable}.
%
In terms of the element-wise lower and upper bounds, namely $\bar{l}_{2,j} \leq \bar{z}_{2,j} \leq \bar{u}_{2,j}$, that pertain to the maxpool functions, each maxpool function is decomposed into multiple ReLU activations (cf. \eqref{eq:toy_ex_maxpool relaxation}). Thus computing these bounds layer-by-layer as in \citet{wong2018provable} would be very costly. The values for $\bar{l}_{2,j}$ and $\bar{u}_{2,j}$ can be calculated more efficiently as follows: For (cf. \eqref{toy_ex_maxpool seperate - a})
%
\begin{equation*}
\bar{z}_{2,j} = z^R_{2,j} - z^M_{2,j},
\end{equation*}
if the element-wise lower and upper bounds for $z^R_{2,j}$ and $z^M_{2,j}$, namely
%
\begin{gather*}
l^R_{2,j} \leq z^R_{2,j} \leq u^R_{2,j}\quad \text{and} \quad
l^M_{2,j} \leq z^M_{2,j} \leq u^M_{2,j}
\end{gather*}
are known, then the element-wise lower and upper bounds $\bar{l}_{2,j}$ and $\bar{u}_{2,j}$ for ${\bar z}$ are derived as
%
\begin{gather*}
\bar{l}_{2,j} = l^R_{2,j} - u^M_{2,j} \quad \text{and} \quad
\bar{u}_{2,j} = u^R_{2,j} - l^M_{2,j}.
\end{gather*}
The elementwise bounds $l^R_{2,j}$ and $u^R_{2,j}$ on the pre-maxpool activations can be computed in a way similar to how the elementwise bounds for pre-ReLU activations are computed in \citet{wong2018provable}. To compute $l^M_{2,j+1}$ and $u^M_{2,j+1}$, recall that
%
%
\begin{equation*}
\begin{aligned}
    &z^M_{2,j+1} = \max\left\{ z^R_{2,j'}:  j' \in \llbracket 0, j \rrbracket \right\}. \  \\
\end{aligned}
\end{equation*}
Thus, one can take
\begin{equation*}
    l^M_{2,j+1} = \max\left\{ l^R_{2,j'}:  j' \in \llbracket 0, j \rrbracket \right\},\quad  \text{and} \quad
    u^M_{2,j+1} = \max\left\{ u^R_{2,j'}:  j' \in \llbracket 0, j \rrbracket \right\}. 
\end{equation*}
Since our algorithm design is primarily an extension of the method proposed by \citet{wong2018provable}, we can deduce the time complexity of our algorithm to be $O(W^2NK)$ by analyzing that the time complexity of \citet{wong2018provable}'s algorithm is $O(W^2N)$.

\begin{table}[H]
\caption{Network parameters} \label{tab: network parameters}
\begin{center}
\begin{tabular}{ |c|c|c|c| }
\hline
    Dataset        &   Model                                                                                            & \# Hidden layers &   \# Parameters \\
 \hline 
 \hline
             ~ & convSmall & 3 & 89606\\
             MNIST & convMed & 3 & 160070\\
             ~ & convBig & 7 & 893418\\
\hline
             ~ & convSmall & 3 & 125318 \\
             CIFAR10 & convMed & 3 & 208198\\
             ~ & convBig & 7 & 2466858\\
 \hline
              ~ & conv$_S$ & 3 & 9538\\
             MNIST & conv$_M$ & 4 & 19162\\
             ~ & conv$_L$ & 4 & 568426\\
\hline
\end{tabular}
\end{center}
\end{table}
\begin{table*}[ht]
\centering
\caption{{Comparison with previous studies using the MNIST dataset}}
    \label{tab: MNIST table 1}
\begin{tabular}{ |c|c|c|c|c|c|c|c| }
\hline
      Model & Training & Accuracy & $\epsilon$ & \makecell[c]{Our \\ Ver}  & \makecell[c]{DeepZ \\ Ver}  & \makecell[c]{DeepPoly \\ Ver}  & \makecell[c]{PRIMA \\ Ver} \\
 \hline 
 \hline
     convSmall & Normal & 100 & 0.03 & 65  & 0  & 40  & \textbf{79}  \\
     convSmall & Fast & 100 & 0.03 & 85  & 2 & 72  & \textbf{92} \\
     convSmall & PGD & 100 & 0.03 & \textbf{97}  & 10  & 91  & \textbf{97} \\
     convSmall & PGD & 100 & 0.04 & 84  & 0  & 53  & \textbf{85} \\
     convMed & Normal & 100 & 0.01 & 93  & 89  & 95  & \textbf{96} \\
     convMed & Fast & 100 & 0.03 & 88  & 49  & 88  & \textbf{96} \\
     convMed & PGD & 100 & 0.03 & 96  & 82  & 96  & \textbf{97} \\
     convMed & PGD & 100 & 0.04 & 89  & 15  & 87  & \textbf{93} \\
     convBig & Normal & 100 & 0.01 & 73  & 0  & 86  & \textbf{86} \\
 \hline
\hline
      Model & Training & Accuracy & $\epsilon$ & \makecell[c]{Our \\ Time} & \makecell[c]{DeepZ \\ Time} & \makecell[c]{DeepPoly \\ Time} & \makecell[c]{PRIMA \\ Time} \\
 \hline 
 \hline
     convSmall & Normal & 100 & 0.03 & \textbf{1.14} & 3.76 & 9.47 & 229.8 \\
     convSmall & Fast & 100 & 0.03 & \textbf{1.17} & 4.55 & 8.95 & 71.28 \\
     convSmall & PGD & 100 & 0.03 & \textbf{1.23} & 3.68 & 6.82 & 25.35 \\
     convSmall & PGD & 100 & 0.04 & \textbf{1.95} & 3.90 & 8.19 & 149.7 \\
     convMed & Normal & 100 & 0.01 & \textbf{4.67} & 5.94 & 10.74 & 15.39 \\
     convMed & Fast & 100 & 0.03 & \textbf{4.24} & 5.60 & 10.11 & 35.16 \\
     convMed & PGD & 100 & 0.03 & \textbf{4.72} & 4.73 & 9.20 & 26.98 \\
     convMed & PGD & 100 & 0.04 & \textbf{4.56} & 5.88 & 10.93 & 58.77 \\
     convBig & Normal & 100 & 0.01 & 338 & \textbf{120.1} & 1493 & 5560 \\
 \hline
\end{tabular}
\end{table*}

\begin{table*}[ht]
\centering
\caption{Comparison with $\alpha,\beta$-CROWN using the MNIST dataset}
    \label{tab: MNIST table 2}
\begin{tabular}{ |c|c|c|c|c|c| }
\hline
      Model & Training & Accuracy & $\epsilon$ & \makecell[c]{Our \\ Ver} &  \makecell[c]{$\alpha\text{-}\beta\text{-crown}$ \\ Ver} \\
 \hline 
 \hline
     conv$_S$ & Normal & 98 & 0.03 & 66 & 0 \\
     conv$_S$ & Fast & 99 & 0.03 & 94 & 33 \\
     conv$_S$ & PGD & 99 & 0.03 & 95 & 37 \\
     conv$_S$ & PGD & 99 & 0.04 & 93 & 11 \\
     conv$_M$ & Normal & 99 & 0.01 & 97 & 13 \\
     conv$_M$ & Fast & 98 & 0.03 & 95 & 13 \\
     conv$_M$ & PGD & 100 & 0.03 & 96 & 7 \\
     conv$_M$ & PGD & 100 & 0.04 & 90 & 2 \\
     conv$_L$ & Normal & 99 & 0.01 & 95 & 0 \\
 \hline
\end{tabular}
\end{table*}

\section{Experiments}\label{sec:experiment_main}
Our experiments compare CAPM with other neural network verification methods, including DeepZ, DeepPoly, PRIMA, and $\alpha,\beta$-CROWN, under various \(\ell_\infty\)-norm perturbation budgets and adversarial attacks (FGSM, PGD). Experiments use CNN architectures (convSmall, convMed, convBig, etc. as shown in Table~\ref{tab: network parameters}) trained on MNIST and CIFAR10, both normally and with adversarial training (Fast and PGD). Verified robustness (abbreviated as Ver.) is defined as the fraction of correctly classified images proven robust against adversarial perturbations, while average verification time (abbreviated as Time) gauges computational cost. The details of experiment can be found in Sec.~\ref{section:experiment}

The results of MNIST are shown in Table~\ref{tab: MNIST table 1} and ~\ref{tab: MNIST table 2}. Based on these tables, CAPM achieves comparable or superior verified robustness to state-of-the-art methods at significantly lower runtime. Although PRIMA achieves the best verified robustness, however, its computational cost is much larger than CAPM and grows with larger perturbation budgets \(\epsilon\). On the other hand, CAPM’s runtime remains nearly constant across all \(\epsilon\). 
$\alpha,\beta$-CROWN faces memory and speed issues on maxpool-based CNNs, especially when it must resort to branch-and-bound (BaB). More experiment result, including the result of CIFAR, can be found in Sec.~\ref{subsec:experiment result}. Overall, CAPM emerges as a robust and efficient solution, particularly well-suited to large-scale verification tasks.

\section{Conclusion}
\label{section:conclusion}

This study {extends \citet{wong2018provable}'s work} to general purpose CNNs with maxpool, padding, and striding operations. {The key idea for handing the maxpool function is to decompose it into multiple ReLU functions, while special care is taken to speed-up the computation of element-wise bounds required for the convex relaxation of intermediate ReLUs in maxpool layer.} General purpose CNNs are expressed using a dual network, which allows efficient computation of verified bounds for CNNs.



The experimental results show that CAPM outperforms previous methods (DeepZ, DeepPoly, and PRIMA) in terms of verified robustness and computational cost for most adversary budget settings, and especially for large-scale CNNs for color images. For an adversary budget $\epsilon=0.0024$, the verified robustness for DeepZ, DeepPoly and PRIMA for convSmall CIFAR10 decreases by to 1\%, 25\%, and 26\%, respectively, but CAPM has a verified robustness of 87.5\%; For an adversary budget $\epsilon = 0.0006$ for convBig CIFAR10, CAPM is 40-times and 20-times faster than PRIMA and DeepPoly, respectively, and gives a significantly higher verified robustness (see Fig.~\ref{fig:compare cifar10} and Fig.~\ref{fig:time cifar10}).


The proposed method gives comparable or better verification with significantly less runtime cost. Unlike many verification methods, for which runtime increases with the adversary budget $\epsilon$, CAPM has a constant runtime, regardless of the adversary budget, so it can be used for larger-scale CNNs which are usually computationally prohibitive for other verification methods. The proposed verification method is suited for use with large scale CNNs, which are an important element of machine learning services. 

%

This study does provide a more precise and efficient verification for maxpool-based CNNs but the verified network is limited to a specific architecture that is defined in Supplementary Material \ref{subsubsec:Notation and data flow}. Future study will involve the design of a verification framework that is applicable to neural networks with a more flexible architecture. Additionally, the method of simplifying certain layers into multiple ReLU layers may likely be limited to maxpool layers only. Therefore, our preliminary future direction will focus on achieving greater flexibility in maxpool-based CNNs. Subsequently, we will continue exploring more flexible neural network architectures, such as residual connections.


\section*{Acknowledgments}
This work was supported in part by the GCP credit award from Google Cloud, the Asian Office of Aerospace Research \& Development (AOARD) under Grant NTU-112HT911020,
National Science and Technology Council of Taiwan under Grant NSTC-112-2221-E-002-204- and NSTC-112-2622-8-002-022 and NSTC-113-2221-E-002-208-,
Ministry of Education (MOE) of Taiwan under Grant NTU-113L891406,
Ministry of Environment under Grant NTU-113BT911001, and ASUS under Project 112CB244.

\bibliography{tmlr}
\bibliographystyle{iclr2025_conference}

\newpage
\appendix
\section{Appendix}
\subsection{Experiment Details}
\label{section:experiment}
This section determines the verified robustness and the average verification time for CAPM, DeepZ, DeepPoly, PRIMA \citep{muller2022prima} and $\alpha$,$\beta$-CROWN for a $l_\infty$ norm-bounded perturbation of various budgets and for various attack schemes, such as FGSM and PGD. Sec.~\ref{subsec:experiment setting} details the experimental network architecture and the input dataset. Sec.~\ref{subsec:experiment result} compares the results for this study with those of previous studies (DeepZ, DeepPoly, PRIMA and $\alpha$,$\beta$-CROWN) in terms of the verified robustness and the average verification time, and Supplementary Material~\ref{sec:reproduce_sota} illustrates how we reproduced the state-of-the-art methods so that we can have a fair comparison with them. Experiments in Supplementary Materials~\ref{subsubsec:Verify the idea of simulating polytope} also demonstrate that the neural network verification problem cannot be simply evaluated using a Monte-Carlo simulation. 
All experiments were conducted on a 2.6 GHz 14 core Intel(R) Xeon(R) CPU E5-2690 v4 with a 512 GB main memory.
%
\subsubsection{Experiment setting}\label{subsec:experiment setting}




\paragraph{Benchmarks:} 
Robustness is calculated against adversarial examples on several networks that are trained using different methods:
%
\begin{itemize}
    \item \textbf{Dataset:} 
    Models are trained using the MNIST and CIFAR10 datasets. Images are normalized using the default setting for DeepPoly \citep{singh2019abstract}. For MNIST, the mean and standard deviation is $0.5$ and $0.5$, respectively. For CIFAR 10, the mean and standard deviation of the RGB channels is $(0.485, 0.456, 0.406)$ and $(0.229, 0.224, 0.225)$, respectively.
    %
    %
    \item \textbf{Architecture of neural networks:}
    There are no empirical verificaion results on maxpool-based CNNs so maxpool layers are added to the common benchmark networks, convSmall, convMed and convBig in \citet{mirman2018differentiable}. The parameter for striding and padding is adjusted to achieve a similar number of parameters to previous studies. Information about these 6 networks is shown in Table \ref{tab: network parameters}. The detail structures are shown in Supplementary Material \ref{appendices: network structure}. \yw{Moreover, although the authors of $\alpha$,$\beta$-CROWN didn't conduct experiments in maxpool-based CNNs before and they did make some additional assumptions on their maxpool layers, we would like to compare with them in maxpool-based CNNs. Hence, we also create the network benchmark conv$_S$, conv$_M$ and conv$_L$ for their settings.}
    %
    %
    \item \textbf{Training methods:} 
    %
    We compared the verification results of CNNs trained either normally (without adversarial training) or with adversarial examples such as Fast-Adversarial \citep{wong2020fast} and PGD \citep{madry2017towards}.
    \item \textbf{Performance metrics:}
    %
    The performance of neural network verification is often evaluated through the following metrics \citep{singh2019abstract}:
        \begin{itemize}
            \item \textit{Verified robustness:} 
            This is  expressed as the number of images verified to be resilient to adversary example attack, divided by the total number of accurate images. This ratio represents the analysis precision of a verifier when a neural network is applied to a test image dataset that {is subject to} attack by an adversarial example.
            %
            %
            \item \textit{Average verified time:} 
            This is the total time that is required by the verification algorithm to verify images, divided by the total number of images.
            %
            %
        \end{itemize}
\end{itemize}


\paragraph{Robustness evaluation:}
For each test dataset, the settings for DeepPoly \citep{singh2019abstract} are used and the top $100$ clean images are used as the evaluation test dataset. Adversarial examples are generated by adding to clean images with $l_\infty$ norm-bounded perturbation for various budgets $\epsilon$ and various attack schemes, such as FGSM and PGD. The generated adversarial examples are then applied to the neural network to compare the accuracy of lower bounds that are evaluated using various verification methods. A {better} verification method must give a tighter (higher) accuracy for the lower bound and never exceeds that of existing attack schemes. The implementation details for DeepPoly, PRIMA and $\alpha$,$\beta$-CROWN are described as follows:
%

%
\begin{itemize}
    \item \textbf{DeepZono and DeepPoly}: 
    %
    We follow the implementation as suggested by the default command in their GitHub \citep{eth-sri}.
    \item \textbf{PRIMA}: 
    %
    We use exactly the same configuration mentioned in \citet{muller2022prima} for convSmall and convBig. Since PRIMA didn't report any results on convMed, and that convMed has the same number of layers as convSmall, we use the same configuration that PRIMA applied to convSmall on convMed as well. 
    \item \textbf{$\alpha$,$\beta$-CROWN}: 
    \yw{
    We used the default parameters and set the parameter conv mode to matrix in order to enable the operation of the BaB algorithm. The examples that would trigger the BaB algorithm were considered not verified. This adjustment was made because the implementation of the BaB algorithm on maxpool-based CNNs results in excessive GPU memory requirements (more than 13GB) and extended execution times (more than 5 minutes per example).
    }
\end{itemize}

\subsubsection{Experimental results}\label{subsec:experiment result}

\begin{figure*}
    \includegraphics[width=1.0\textwidth]{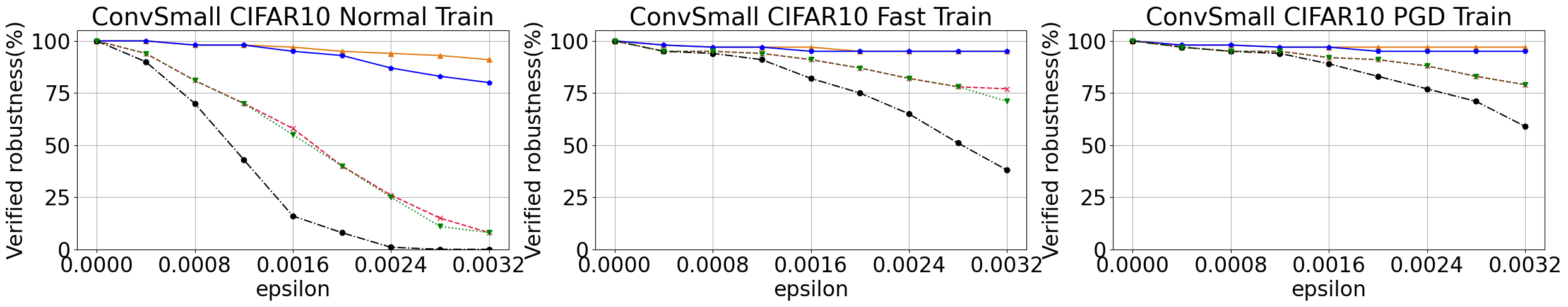}
    \includegraphics[width=1.0\textwidth]{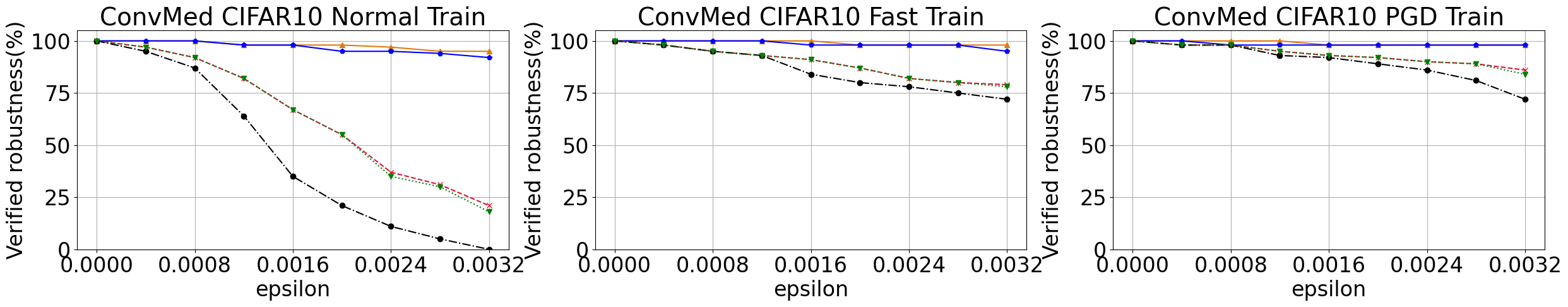}
    \includegraphics[width=1.0\textwidth]{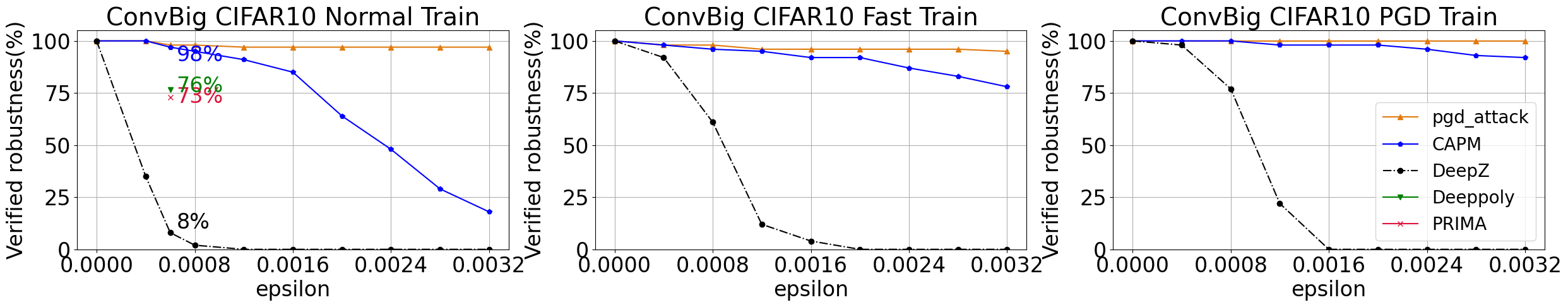}
    \caption{
    Verified robustness for $\epsilon$ perturbations under $l_\infty$-norm by CAPM (black solid line with pentagon marker), DeepPoly (green dotted line with triangle marker), DeepZ (black dashdot line with circle
    marker), and PRIMA (red dashed line with x marker) for convSmall CIFAR10, convMed CIFAR10, and convBig CIFAR10. The orange solid line with triangle markers is the classification accuracy for a PGD attack. PRIMA and DeepPoly are both prohibitively computationally costly for convBig CIFAR10, so we only show the result for DeepZ.
    }
    \label{fig:compare cifar10}
\end{figure*}

\begin{figure*}
    \includegraphics[width=1.0\textwidth]{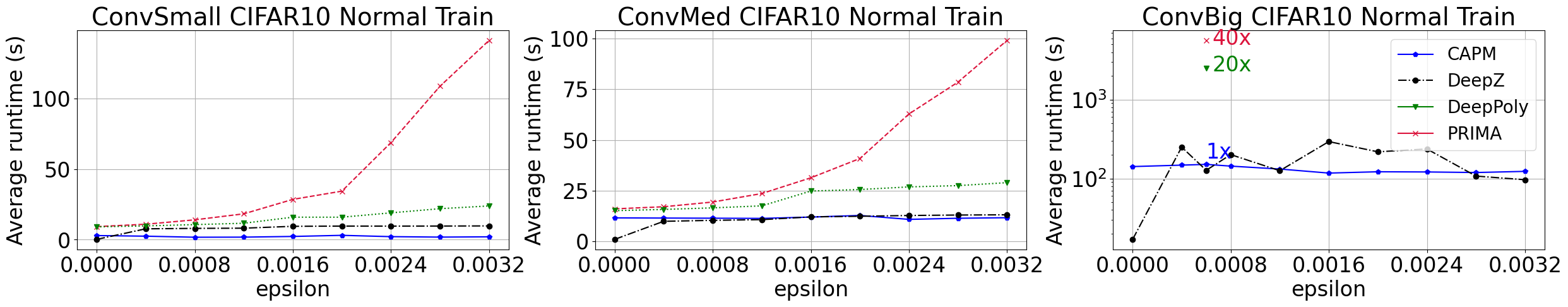}
    \caption{
    Runtime for $\epsilon$ perturbations under $l_\infty$-norm  for CAPM (black solid line with pentagon marker), DeepPoly (green dotted line with triangle marker), DeepZ (black dashdot line with circle marker) and PRIMA (red dashed line with x marker) for convSmall CIFAR10, convMed CIFAR10 and convBig CIFAR10.
    }
    \label{fig:time cifar10}
\end{figure*}

The results for CAPM are compared with those of previous studies (DeepZ, DeepPoly, PRIMA) in terms of the verified robustness and average runtime metrics for various adversary budgets $\epsilon$ for six different networks, as shown in Fig.~\ref{fig:compare cifar10} and Fig.~\ref{fig:time cifar10}. The classification accuracy
\footnote{
Here we follow the settings in \citet{singh2019abstract,muller2022prima} and only test on images which were correctly classified before any perturbation is added.}
for a real-world PGD attack (orange solid line with triangle marker) is also determined. A verification method must demonstrate verified robustness that is no greater than the accuracy of real-world attack schemes. 
Fig.~\ref{fig:compare cifar10} and Fig.~\ref{fig:time cifar10} illustrate that CAPM achieves better verification robustness than all other schemes for all combination of the CIFAR10 dataset and has a much lower computational cost. The computational cost of CAPM is independent of $\epsilon$, because a different adversary budget corresponds to the same dual network architecture so computational costs are similar. However, the verification time that is required by PRIMA increases as $\epsilon$ increases, possibly because as $\epsilon$ increases, the intermediate adversarial polytope becomes more complicated and must be described by a more complex MILP optimization problem. This demonstrates the promising potential of CAPM towards verification of large scale CNNs and colour images. \yw{Due to the excessive GPU memory requirements of $\alpha$,$\beta$-CROWN, we did not include $\alpha$,$\beta$-CROWN in this set of experiments.}

\begin{figure}[ht]
\centering
\includegraphics[width=0.7\textwidth]{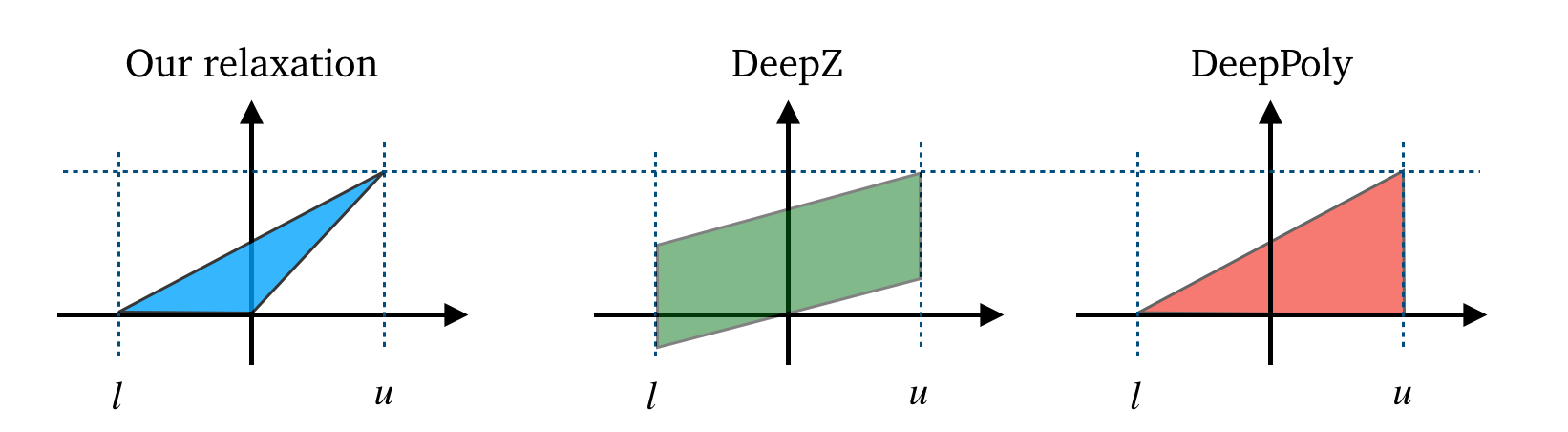}
\caption{\label{fig:diff relaxation}
The difference in convex relaxation for the method of this study, DeepZ, and DeepPoly.}
%
\end{figure}

The results for the MNIST dataset are shown in Table \ref{tab: MNIST table 1} and Table \ref{tab: MNIST table 2}. PRIMA and CAPM both yield a higher (tighter) verification robustness than DeepPoly or DeepZ (cf. Fig.~\ref{fig:diff relaxation}). CAPM also has a comparable verification robustness to PRIMA, but average runtime is significantly reduced. For larger networks such as convBig, only CAPM achieves effective verified robustness in a reasonable runtime.
%

\yw{ 
For $\alpha$,$\beta$-CROWN, the BaB algorithm they employ is not suitable for maxpool CNNs, leading to excessive GPU memory requirements (more than 13GB) and prolonged execution times (more than 5 minutes per example). Therefore, in this experiment, we considered examples that would trigger the BaB algorithm as not verified. Consequently, the performance of $\alpha$,$\beta$-CROWN is not satisfactory in our study.
}

CAPM has the following advantages, compared to other state of the arts verification methods (DeepZ, DeepPoly, and PRIMA):
%

\begin{itemize}
    \item 
    CAPM achieves comparable or better verification robustness with significantly less runtime cost.
    %
    %
    \item 
    Unlike other verification methods for which runtime increases with the adversary budget $\epsilon$, CAPM has a constant runtime that is independent of $\epsilon$. For larger networks such as convBig, only CAPM demonstrates a feasible runtime and good verified robustness.
    %
    %
    \item 
    CAPM is especially suitable for larger-scale maxpool-based CNNs that are designed for color images. Using the CIFAR10 dataset, CAPM achieves a significantly tighter verified robustness at a much lower computational cost. 
    %
    \yw{And due to computational costs, we did not continue our experiments on larger and more general networks. Nevertheless, from Tables~\ref{tab: MNIST table 1} and~\ref{tab: MNIST table 2} in our experiments, it can be observed that our CAPM not only exhibits a lower growth rate in computation time as the neural network scales up but also maintains nearly equal accuracy under significantly lower computation time compared to PRIMA.}
\end{itemize}

%


\subsection{Maxpool-based CNN architecture}\label{subsec:Maxpool-based CNN Architecture}
As mentioned in section \ywz{\ref{subsec:Verification Problem}}, the network $f$ in consideration is a maxpool-based CNN, which consists of $k$ convolution layers with ReLU activation and maxpool functions, followed by flattening and $t$ fully connected layers with ReLU activation. In this manuscript we separate layers $l-1$ and $l$ with the linear operations.  That is, the first layer simply contains a convolution operation; and from layer $2$ to $k-1$, there are ReLU, maxpool, and convolution operations in order; while the $k$-th layer contains ReLU, maxpool, flatten, and linear (fully connected) operations. After the flatten operation is the fully-connected part of the network. Denote $T = k+t$ as the total number of layers in $f$, then within layers $k+1$ to $T$ there are ReLU and linear (fully connected) operations in order, while the last layer is a pure output layer containing no operations.  
In the following derivation, we consider CNN with $k \geq 2$ and $t \geq 1$. 
The explicit notations pertaining to all intermediate results during the calculation of $f$ are specified in Sec.~\ref{subsubsec:Notation and data flow} and Table \ref{tab:basic_notation}.

\begin{table*}[h]
\centering
\caption{\label{tab:basic_notation} Basic notations, where $\llbracket m,n \rrbracket$ denotes the set of integers from $m$ to $n$.}
\begin{tabular}{ |l|c|l|} 
 \hline
    layer & notation & description \\
 \hline
    $l=1$   & $\mathbf{z}_l$         &  the (perturbed) input image. \\
 \hline
    $l \in \llbracket 1,k \rrbracket$   & $c_l$       & the channel number of the $l$-th layer feature maps.\\
         \        & $N_l$       & the width and height of feature map $\mathbf{z}_l$. \\
 \hline
    $l \in \llbracket 1,k-1\rrbracket$ & $k_l^{cv}$ & the size of the $l$-th convolution kernel. \\
         \        & $s_l^{cv}$ & the stride of the $l$-th convolution kernel.\\
         \        & $p_l^{cv}$ &  the padding of the $l$-th convolution kernel (choose 0 or greater).  \\
    \hline
    $l \in \llbracket 2,k \rrbracket$   & $\hat{\mathbf{z}}_l$&  the feature map after convolution which will be fed to ReLU.  \\
         \        & $\mathbf{z}^R_l$    &  the feature map after ReLU which will be fed to maxpool.\\
         \        & $\mathbf{z}_l$  & the feature map after maxpool which will be fed to convolution.\\
         \        & $k_l^M$    &  the size of the $l$-th maxpool kernel.\\
         \        & $s_l^M$    &  the size of the $l$-th maxpool stride.\\
         \        & {$q_l$}      &  the width and height of feature maps                               $\hat{\bm{z}}_l$ and $\bm{z}^R_l$. \\
 \hline   
  $l \in \llbracket k+1,k+t \rrbracket$& $\hat{\mathbf{z}}'_l$  &  the feature vector after (fully connected) linear operator in layer \\
         \        & \          & $l-1$, which will be fed to ReLU.\\
 \hline
  $l \in \llbracket k,k+t-1\rrbracket$ & $\Tilde{\mathbf{z}}_l$ &  the feature vector before (fully connected) linear operator in layer $l$. \\
 \hline
  $l \in \llbracket k,k+t \rrbracket$ & $a_l$      &  the length of vectors $\hat{\mathbf{z}}'_l$ and $\Tilde{\mathbf{z}}_l$. \\
  \hline 
\end{tabular}
\end{table*}

\begin{figure*}[h]
\centering
\includegraphics[width=1.0\textwidth]{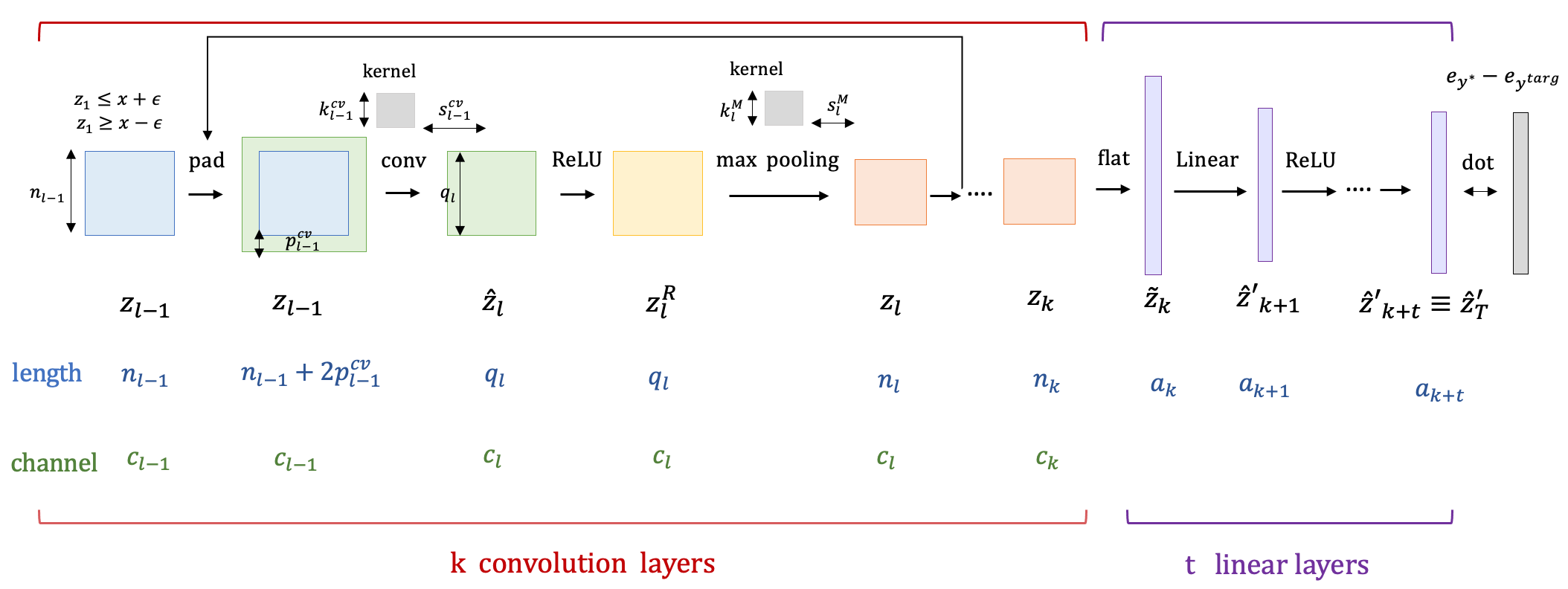}
\caption{\label{fig:basic_notation.png} Data flow and notations.}
\end{figure*}

\subsubsection{Notation and data flow}
\label{subsubsec:Notation and data flow}
We use the tuple $(l,c,m,n)$ to indicate the pixel located at the $m$-th row and $n$-th column in the $c$-th channel of the corresponding $l$-th layer feature map. For instance, such a pixel in the feature map $\mathbf{z}_l$ is indicated as $z_{l,c,m,n}$.

\begin{figure}[ht]
\centering
\includegraphics[width=0.2\textwidth]{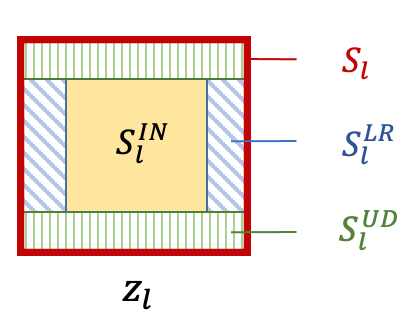}
\caption{\label{fig:S_l.png} Pixel regions of feature map $\mathbf{z}_l$.}
\end{figure}

To consider padding which is a common practice in convolutions, we extend the pixel region of the feature map $\mathbf{z}_l$, which is the input to the convolution operation, to $S_l = S_l^{IN} \cup S_l^{UD} \cup S_l^{LR}$ as indicated in Table \ref{tab:The definition of region of map} and Fig.~\ref{fig:S_l.png}. Here $S_l^{IN}$ represents the index region of the feature map $\mathbf{z}_l$ without padding, while $S_l^{UD}$ represents the padded index region located at the upper/down-sides of the feature map $\mathbf{z}_l$, and $S_l^{LR}$ represents the padded index region located at the left/right-sides of the feature map $\mathbf{z}_l$.

Refer to Fig.~\ref{fig:basic_notation.png}, in the convolution part of the network, which corresponds to layers $l \in $ $\llbracket 1,k \rrbracket$, we first compute feature map $\hat{\mathbf{z}}_l$ as the convolution of $\mathbf{z}_{l-1}$. Starting from $\hat{\mathbf{z}}_l$, we consecutively apply ReLU and maxpool to compute intermediate results $\mathbf{z}^R_l$ and $\mathbf{z}_l$, respectively. Note that $\hat{\mathbf{z}}_l$ and $\mathbf{z}^R_l$ both have the same index region ${Q_l^{IN}}$ (cf. Table \ref{tab:The definition of region of map}) as ReLU does not change the size of feature map. The aforementioned process can be iterated until reaching $\mathbf{z}_k$, to which the flat kernel is applied to flatten the feature map $\mathbf{z}_k$ into vector form $\Tilde{\mathbf{z}}_k$. 
In the fully-connected part of the network, which corresponds to layers $l \in$ $\llbracket k+1,k+t \rrbracket$, we first compute feature vector $\hat{\mathbf{z}}'_l$ as the result of the fully connected linear operation applied on $\Tilde{\mathbf{z}}_{l-1}$, followed by ReLU to get $\Tilde{\mathbf{z}}_l$. We iterate the aforementioned process until finally reaching $\hat{\mathbf{z}}'_T$, which is the output of network $f$.

\begin{table*}
    \centering
    \caption{Fixed region notations.}
    \begin{tabular}{|l|l|}
        \hline
         layer         & notation description  \\
        \hline
         $l = 1,...,k$ & $S_l^{IN} = \llbracket 0,c_l-1\rrbracket \times \llbracket 0,N_l-1 \rrbracket \times \llbracket 0,N_l-1\rrbracket$ \\
        \hline
         $l=1,...,k-1$ & $S_l^{UD} = \llbracket 0,c_l-1 \rrbracket \times (\llbracket -p_l^{cv},-1\rrbracket \cup \llbracket N_l,N_l+p_l^{cv}-1 \rrbracket) \times \llbracket -p_l^{cv},N_l+p_l^{cv}-1 \rrbracket$ \\
                \      & $S_l^{LR} = \llbracket 0,c_l-1\rrbracket \times \llbracket 0,N_l-1\rrbracket \times (\llbracket -p_l^{cv},-1 \rrbracket \cup \llbracket N_l,N_l+p_l^{cv}-1 \rrbracket)$ \\
                \      & $S_l = S_l^{IN} \cup S_l^{UD} \cup S_l^{LR}$ \\
        \hline
        $l=2,...,k$  & $Q_l^{IN} = \llbracket 0,c_l-1\rrbracket \times \llbracket 0,{ q_l}-1 \rrbracket \times \llbracket 0,{ q_l}-1 \rrbracket$ \\
        \hline
    \end{tabular}
    \label{tab:The definition of region of map}
\end{table*}

{
Before introducing the whole process of network $f$, we first introduce notations for the convolution kernel and the flat kernel as below:\\
}

\begin{figure}[H]
\centering
\includegraphics[width=0.5\textwidth]{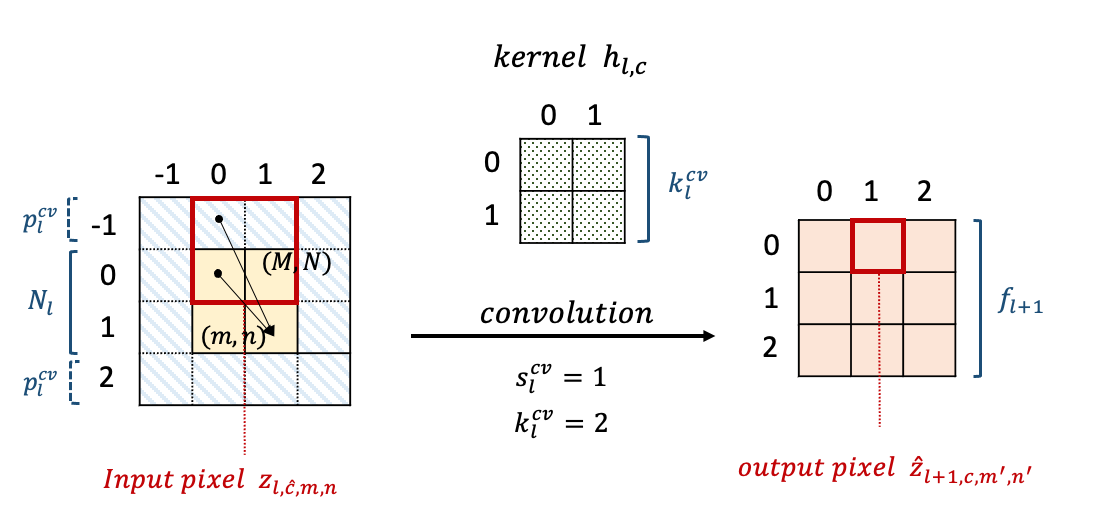}
\caption{\label{fig:valid kernel index.png} 
    {
    After convolution the output pixel $\hat{z}_{l+1,c,m',n'}$ is the sum of products between $z_{l,\hat{c},m,n}$ and $h_{l,c}(\hat{c},M,N)$, where
    $M=m+p_l^{cv}-s_l^{cv}m'$ and $N=n+p_l^{cv}-s_l^{cv}n'$.
    }
}
\end{figure}

\begin{itemize}
    \item Convolution kernel $h_{l,c}$: As illustrated in Fig.~\ref{fig:valid kernel index.png}, we use $h_{l,c}$ to represent the $c$-th convolution kernel in the $l$-th layer, where $l \in \llbracket 1, k-1 \rrbracket$ and $c \in \llbracket 0,c_{l+1}-1 \rrbracket$. In addition, each kernel contains $c_l$ channels with both height and width $k_l^{cv}$, namely the index region of $h_{l,c}$ is $\llbracket0,c_l-1\rrbracket \times \llbracket0,k_l^{cv}-1\rrbracket \times \llbracket0,k_l^{cv}-1\rrbracket$. Here we use $h_{l,c}(\hat{c},\hat{m},\hat{n})$ to refer to the kernel value at the $\hat{c}$-th channel, $\hat{m}$-th row and $\hat{n}$-th column, and define $h_{l,c}(\hat{c},\hat{m},\hat{n}) = 0$ for each index $(\hat{c},\hat{m},\hat{n})$ that is out of the range $\llbracket0,c_l-1\rrbracket \times \llbracket0,k_l^{cv}-1\rrbracket \times \llbracket0,k_l^{cv}-1\rrbracket$.
    \item Flat kernel $\hat{\bm{W}}$: The flat kernel unfolds the feature map $\bm{z}_k$ to the vector $\Tilde{\bm{z}}_k$ in the flattening layer, namely 
    $$\Tilde{z}_{k,a} = \sum\limits_{(c,m,n) \in S_k^{IN}} (\hat{W}_{c,m,n})_a z_{k,c,m,n}, ~~ a \in \llbracket 0, a_k-1 \rrbracket,$$ 
    where $(\hat{W}_{c,m,n})_a =\mathbbm{1}\{a = c_kN_k N_k + mN_k + n\}$ for $(c,m,n) \in S_k^{IN}$, and that $a_k = c_kN_kN_k$.
\end{itemize}
Before introducing the details of the math operations processing through network $f$, we first illustrate how to decompose the maxpool: Instead of considering the whole group of candidates as the input of maxpool, e.g., $y_{out} = \max(x_1,x_2,...,x_N)$, we split the maxpool function into several lines of one-by-one comparisons, each considering only one additional input
\begin{equation} \label{eq:maxpool rough idea}
    \begin{aligned}
        y_1 &= \max(x_1,0) = \mbox{ReLU}(x_1-0)+0, \\
        y_2 &= \max(x_2,y_1) = \mbox{ReLU}(x_2-y_1) + y_1, \\
        & \vdots \\
        y_{out} &= \max(x_N,y_{N-1}) = \mbox{ReLU}(x_N-y_{N-1}) + y_{N-1}.
    \end{aligned}
\end{equation}
Note that we assume all of the inputs $x_1,...,x_N$ are non-negative, since they are drawn from the output of ReLU. The operations through network $f$ is given as follows:

Starting with feature map $\bm{z}_{l-1}$ ($l \in \llbracket 2, k \rrbracket$) with index region $S^{IN}_{l-1}$, to apply convolution with zero-padding, we first pad the four boundaries of $\bm{z}_{l-1}$ with $0$, namely
\begin{equation}\label{eq:pad}
     z_{l-1,c,m,n} = 0,\ { (c,m,n)\in S_{l-1}^{UD} \cup S_{l-1}^{LR}},
\end{equation}
\begin{equation}\label{eq:conv}
    \hat{z}_{l,c,m',n'}= b_{l-1,c}     + \sum\limits_{(\hat{c},m,n) \in S_{l-1}} h_{l-1,c}(\hat{c},m+p_{l-1}^{cv}-s_{l-1}^{cv}m',n+p_{l-1}^{cv}-s_{l-1}^{cv}n')z_{l-1,\hat{c},m,n} , (c,m',n')\in Q_{l}^{IN}  \sim  l \in \llbracket 2,k \rrbracket. 
\end{equation}
By taking convolution between the convolution kernel $h_{l-1,c}$ and the padded $\bm{z}_{l-1}$, we compute $\hat{\bm{z}}_l$ as in \eqref{eq:conv}. As such, $\bm{z}^R_{l}$ is obtained by applying ReLU to $\hat{\bm{z}}_{l}$ as follows:
\begin{equation}\label{eq:relu}
    z^R_{l,c,m,n} =\max \Big\{ \hat{z}_{l,c,m,n},  0\Big\}, \ (c,m,n) \in { Q_l^{IN}}. \\ 
\end{equation}
\begin{figure}[H]
\centering
\includegraphics[width=0.4\textwidth]{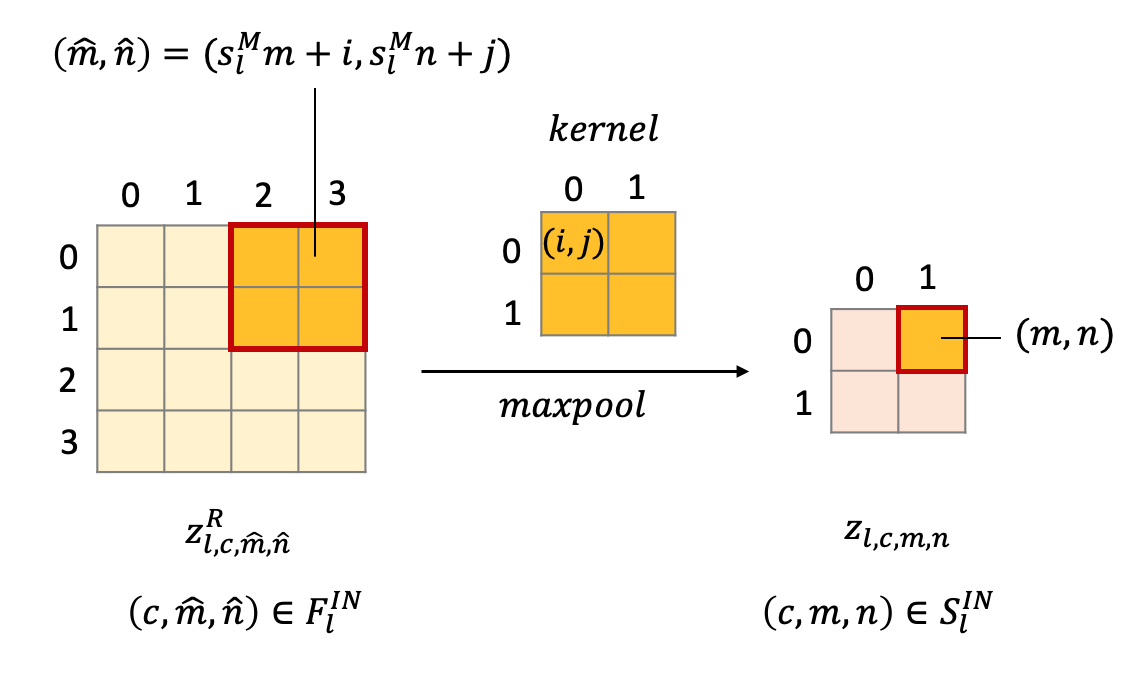}
\caption{\label{fig:maxpool kernel operation} {
{
\small This example demonstrates the relationship between the output pixel $z_{l,c,m,n}$ and input pixels $z^R_{l,c,\hat{m},\hat{n}}$ when applying maxpool function. The input map $z^R_l$ has $1$ channel with  width and height both equal to $4$ (${q_l}=4$), and the kernel size is $2 \times 2$. As such, the pixel $(0,0,1)$ in the output map is the maximum of the input pixels $(0,0,2),(0,0,3),(0,1,2),(0,1,3)$.}\\}}
\end{figure}

Regarding the maxpool layer, as illustrated in Fig.~\ref{fig:maxpool kernel operation}, one has
\begin{equation}\label{eq: maxpool of zR_l and z_l}
    z_{l,c,m,n} = \max\limits_{\substack{i \in \llbracket 0,k_l^M-1 \rrbracket, \\ j \in \llbracket 0,k_l^M-1 \rrbracket }} \Bigg\{  z^R_{l,c,s_l^Mm+i,s_l^Mn+j} 
\Bigg\}
\end{equation}
where $k_l^M$ and $s_l^M$ indicate the kernel size and stride, respectively. By applying the decomposition trick to maxpool as in \eqref{eq:maxpool rough idea}, one may rewrite \eqref{eq: maxpool of zR_l and z_l} as 
$z_{l,c,m,n} = (z^M_{l,c,m,n})_{k_l^Mk_l^M}$ for $(c,m,n) \in S_l^{IN}$, where we denote $(z^M_{l,c,m,n})_0 = 0$ and 
\begin{equation}\label{eq: rewrite maxpool of zR_l and z_l}
\begin{aligned}
    &(z^M_{l,c,m,n})_{ik_l^M+j+1} \\
    &= \max \Big\{ z^R_{l,c,s_l^Mm+i',s_l^Mn+j' } \ :
        \substack{
        i' \in \llbracket 0,k_l^M-1 \rrbracket,\\
        j' \in \llbracket 0, k_l^M-1 \rrbracket, \\ i'k_l^M+j' \leq ik_l^M+j
        }
        \Big\} \\
    &=  \max  \Big\{  z^R_{l,c,s_l^Mm+i,s_l^Mn+j }, \ (z^M_{l,c,m,n})_{ik_l^M+j}  \Big\} \\
    &=   \max \Big\{  z^R_{l,c,s_l^Mm+i,s_l^Mn+j }- (z^M_{l,c,m,n})_{ik_l^M+j} \Big\} + (z^M_{l,c,m,n})_{ik_l^M+j}
\end{aligned}
\end{equation}
for $i \in \llbracket 0 , k_l^M-1 \rrbracket$ and $j \in \llbracket 0, k_l^M-1 \rrbracket$, which can be further decomposed as follows:
\begin{equation} \label{eq:maxpool}
        \left \{
        \begin{aligned}
                &(\Bar{z}_{l,c,m,n})_{ik_l^M+j} = z^R_{l,c,s_l^Mm+i,s_l^Mn+j} - (z^M_{l,c,m,n})_{ik_l^M+j} \\
                &(z'_{l,c,m,n})_{ik_l^M+j} = \quad \max \quad \{(\Bar{z}_{l,c,m,n})_{ik_l^M+j}, 0 \}\\
                &(z^M_{l,c,m,n})_{ik_l^M+j+1} = (z'_{l,c,m,n})_{ik_l^M+j} + (z^M_{l,c,m,n})_{ik_l^M+j} \\
        \end{aligned}
        \right . .
\end{equation}
Once the convolution part of the neural network is completed, we get $z_k$ and convert it into a vector $\Tilde{z}_k$ by flat kernel $\hat{\bm{W}}$, namely,
\begin{equation}\label{eq:flatten}
    \Tilde{z}_{k,a} = \sum\limits_{(c,m,n) \in S_k^{IN}} (\hat{W}_{c,m,n})_a z_{k,c,m,n}
\end{equation}
for all $a \in \llbracket 0, a_k-1 \rrbracket$. Finally, $\Tilde{z}_k$ is fed into linear part, where the linear and ReLU operations can be written as:
\begin{subequations}\label{eq:fully}
    \begin{align}
        \begin{split}
            &\hat{\bm{z}}'_{l+1} = \bm{W}_l\Tilde{\bm{z}}_l + \bm{b}_l, \ l \in \llbracket k, k+t-1 \rrbracket 
        \end{split}\label{eq: linear-linear} \\
        \begin{split}
        & \Tilde{\bm{z}}_l = \quad \max (\hat{\bm{z}}'_{l}, 0), \ l \in \llbracket k+1, k+t-1 \rrbracket. 
        \end{split}\label{eq: linear-ReLU}
    \end{align}
\end{subequations}

\subsubsection{Convex outer bound}\label{subsubsec:Convex Outer Bound}

\begin{figure}[ht]
\centering
\includegraphics[width=0.53\textwidth]{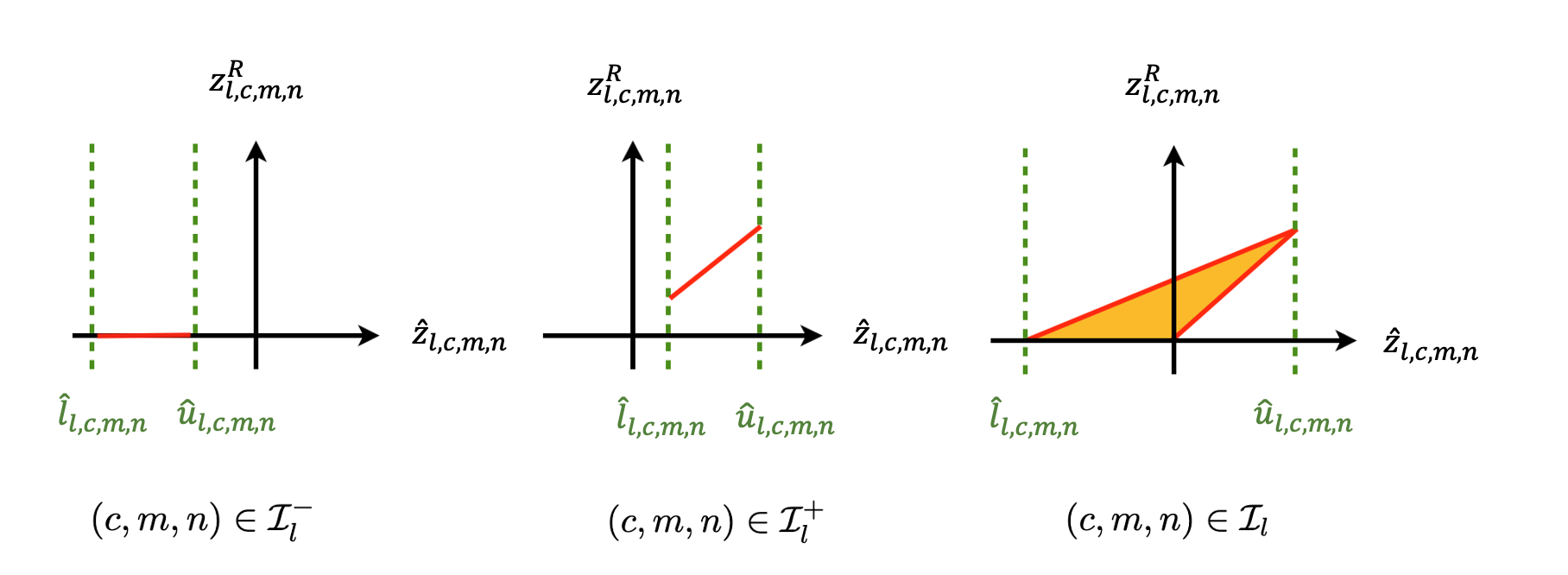}
\caption{\label{fig:convex outer bound.png} {\small Convex outer bound for ReLU function under various scenarios.} }
\end{figure}

In order to make the feasible solution set a convex set, we construct the convex outer bound for the activation function. First, we consider the ReLU functions in network $f$, listed in the following:
\thinmuskip=0mu
\medmuskip=0mu plus 1mu minus 0mu
\thickmuskip=0mu plus 1mu minus 0mu
\begin{subequations}\label{eq: need convex outer bound}
\begin{align}
    & z^R_{l,c,m,n} =\max  (\hat{z}_{l,c,m,n}, 0),  ~~ l \in \llbracket 2,k \rrbracket, (c,m,n) \in Q_l^{IN} \\
    & \begin{array}{l}
    (z'_{l,c,m,n})_{ik_l^M+j} =  \max  ((\Bar{z}_{l,c,m,n})_{ik_l^M+j}, 0), \\
    ~ l \in \llbracket 2,k \rrbracket, (c,m,n) \in S_l^{IN},  (i,j) \in \llbracket 0,k_l^M-1 \rrbracket \times \llbracket 0,k_l^M-1 \rrbracket
    \end{array}\\
    & \Tilde{\bm{z}}_l =\max  (\hat{\bm{z}}'_{l}, \bm{0}), ~~ l \in \llbracket k+1,k+t-1 \rrbracket
\end{align}
\end{subequations}
where $\hat{z}_{l,c,m,n}$,  $(\bar{z}_{l,c,m,n})_{ik_l^M+j}$, and $\hat{\bm{z}}'_l$ are given by \eqref{eq:conv}, \eqref{eq:maxpool} and \eqref{eq: linear-linear}. Suppose for now that we have already known the upper and lower bounds for each entity in the right hand side of \eqref{eq: need convex outer bound} :
 \begin{subequations}\label{eq: upper lower bound of input relu}
     \begin{align}
        \begin{split}
            &\hat{l}_{l,c,m,n} < \hat{z}_{l,c,m,n} < \hat{u}_{l,c,m,n}
        \end{split} \\
        \begin{split}
            & (\bar{l}_{l,c,m,n})_{ik_l^M+j} < (\bar{z}_{l,c,m,n})_{ik_l^M+j} < (\bar{u}_{l,c,m,n})_{ik_l^M+j} \\
        \end{split} \\
        \begin{split}
            &l'_{l,a} < \hat{z}'_{l,a} < u'_{l,a}
        \end{split}\label{upper-lower of hat_z'}
     \end{align}
 \end{subequations}
Following the idea of \citep{wong2018provable}, we categorize the variables $\{\hat{z}_{l,c,m,n}\}$ into three clusters according to the sign of their upper-lower bounds, namely
\begin{equation}
    \begin{aligned}
        \mathcal{I}_l^- &= \{ (c,m,n) \in Q_l^{IN} \ |  \ \hat{l}_{l,c,m,n} \leq \hat{u}_{l,c,m,n} \leq 0 \} \\
        \mathcal{I}_l^+ &= \{ (c,m,n) \in Q_l^{IN} \ |\ 0 < \hat{l}_{l,c,m,n} \leq  \hat{u}_{l,c,m,n}  \} \\
        \mathcal{I}_l &= \{ (c,m,n) \in Q_l^{IN} \ | \ \hat{l}_{l,c,m,n} \leq 0 < \hat{u}_{l,c,m,n}   \} \\
    \end{aligned}
\end{equation}
As illustrated in Fig.~\ref{fig:convex outer bound.png}, for scenarios $ \mathcal{I}_l^-$ and $\mathcal{I}_l^+$, the ReLU function $z^R_{l,c,m,n} = \max (\hat{z}_{l,c,m,n},0)$ is essentially zero function $(z^R_{l,c,m,n}=0)$ and identity function {$(z^R_{l,c,m,n}=\hat{z}_{l,c,m,n})$,} respectively. For scenario $\mathcal{I}_l$, we construct the convex outer bound \citep{wong2018provable} for the ReLU function as follows:
\begin{equation}\label{eq outer bound}
    \begin{aligned}
    &z^R_{l,c,m,n} \geq 0, \\
    &z^R_{l,c,m,n} \geq \hat{z}_{l,c,m,n}\\
    &(\hat{u}_{l,c,m,n}-\hat{l}_{l,c,m,n})z^R_{l,c,m,n} \leq \hat{u}_{l,c,m,n}\hat{z}_{l,c,m,n} -\hat{u}_{l,c,m,n}\hat{l}_{l,c,m,n}\\
    \end{aligned} 
\end{equation}
The convex outer bounds for ReLU functions pertaining to $\Tilde{\bm{z}}_l$ and $(z'_{l,c,m,n})_{ik_l^M+j}$ are constructed analogously.  We cluster the variables $\{\hat{z}'_{l,a}\}$ and ${\{(\bar{z}_{l,c,m,n})_a\}}$ into three different scenarios accordingly:
\begin{equation}
    \begin{aligned}
        \hat{\mathcal{I}}_l^- &= \{ a \in \llbracket 0 , a_{l-1} \rrbracket \ | \ l'_{l,a} \leq u'_{l,a} \leq 0 \} \\
      \hat{\mathcal{I}}_l^+ &= \{ a \in \llbracket 0 , a_{l-1} \rrbracket \ |\ 0 < l'_{l,a} \leq u'_{l,a} \} \\
        \hat{\mathcal{I}}_l &= \{ a \in \llbracket 0 , a_{l-1} \rrbracket \ | \ l'_{l,a} \leq 0 <  u'_{l,a}  \} \\
    \end{aligned}
\end{equation}
\begin{footnotesize}
\begin{equation}\label{eq: three clusters of z bar}
    \begin{aligned}
        \bar{\mathcal{I}}_l^- &= \{  (c,m,n,a) \in S_l^{IN }\times \llbracket 0 , k_l^Mk_l^M-1 \rrbracket \ | \ (\bar{l}_{l,c,m,n})_a \leq (\bar{u}_{l,c,m,n})_a \leq 0 \} \\
      \bar{\mathcal{I}}_l^+ &= \{  (c,m,n,a)  \in S_l^{IN }\times \llbracket 0 , k_l^Mk_l^M-1 \rrbracket \ |\ 0 < (\bar{l}_{l,c,m,n})_a \leq (\bar{u}_{l,c,m,n})_a \} \\
        \bar{\mathcal{I}}_l &= \{  (c,m,n,a)  \in S_l^{IN }\times \llbracket 0 , k_l^Mk_l^M-1 \rrbracket \ | \ (\bar{l}_{l,c,m,n})_a \leq 0 <  (\bar{u}_{l,c,m,n})_a  \}
    \end{aligned}
\end{equation}
\end{footnotesize}

\subsection{Formulation of the Lagrangian dual problem}\label{subsec:Primal problem}

As mentioned earlier, we solve the verification problem with the dual-view approach and take the best dual feasible solution as a lower bound for the verification problem. {The primal and dual problems are given in Sec.~\ref{subsubsec: primal problem} and \ref{subsubsec:Dual Problem}, where the dual problem is further simplified to a dual network in Sec.~\ref{subsubsec:CNN Dual Network}.}

\subsubsection{Primal problem}\label{subsubsec: primal problem}

By summarizing Sec.~\ref{subsec:Maxpool-based CNN Architecture} (cf. \eqref{noise_def},\eqref{objective},\eqref{eq:pad}-\eqref{eq: three clusters of z bar}), the verification problem can be formulated as the  convex optimization problem.

\begin{figure*}
\begin{align*}
        \text{minimize} \quad & \bm{d}^T \hat{\mathbf{z}}'_T \equiv (\mathbf{e}_{y^*}-\mathbf{e}_{y^{targ}})^T \hat{\mathbf{z}}'_T\\
        \textrm{subject to} \quad & \mathbf{z}_1 - (\mathbf{x} + \bm{\epsilon}) \leq \mathbf{0}  \\
& -\mathbf{z}_1 + (\mathbf{x} - \bm{\epsilon}) \leq \mathbf{0}\\
 \quad & z_{l-1,c,m,n} = 0, \ ~(c,m,n)\in S_{l-1}^{UD} \cup S_{l-1}^{LR}, \    { l \in \llbracket 2,k \rrbracket}\\
 \quad &\hat{z}_{l,c,m',n'} - b_{l-1,c} -\sum\limits_{(\hat{c},m,n) \in S_{l-1}} h_{l-1,c}({\hat{c},m+p_{l-1}^{cv}-s_{l-1}^{cv}m',n+p_{l-1}^{cv}-s_{l-1}^{cv}n'})z_{l-1,\hat{c},m,n} = 0, \ ~(c,m',n')\in Q_{l}^{IN} \\
 &~\text{, for } l \in \llbracket 2,k \rrbracket \\
 \quad &\left\{ 
  \begin{aligned}
  & -z^R_{l,c,m,n} = 0 &&, (c,m,n) \in \mathcal{I}_l^- \\
 & \hat{z}_{l,c,m,n} - z^R_{l,c,m,n} = 0 &&,(c,m,n) \in \mathcal{I}_l^+ \\
 & \left .
\begin{aligned}
 & -z^R_{l,c,m,n} \leq 0 \\
& \hat{z}_{l,c,m,n} - z^R_{l,c,m,n} \leq 0 \\
& (\hat{u}_{l,c,m,n} - \hat{l}_{l,c,m,n})z^R_{l,c,m,n} - \hat{u}_{l,c,m,n}\hat{z}_{l,c,m,n} + \hat{u}_{l,c,m,n}\hat{l}_{l,c,m,n} \leq 0 \\
\end{aligned}
\right \} &&,(c,m,n) \in \mathcal{I}_l \\
  \end{aligned} 
  \right \}, \ { l \in \llbracket 2,k \rrbracket} \\
  \quad &\left\{ 
  \begin{aligned}
    &(z^M_{l,c,m,n})_0 = 0 \\
    &(\Bar{z}_{l,c,m,n})_{ik_l^M+j} -  [z^R_{l,c,s_l^Mm+i,s_l^Mn+j} - (z^M_{l,c,m,n})_{ik_l^M+j}] = 0, \ i \in \llbracket 0, k_l^M-1 \rrbracket, \ j \in \llbracket 0, k_l^M-1 \rrbracket, \ (c,m,n) \in S_l^{IN}\\
     & \left \{
      \begin{aligned}
     & -(z'_{l,c,m,n})_a = 0 &&, ~(c,m,n,a) \in \Bar{\mathcal{I}}_l^-\\
     & (\Bar{z}_{l,c,m,n})_a -(z'_{l,c,m,n})_a = 0 &&, ~(c,m,n,a) \in \Bar{\mathcal{I}}_l^+ \\
     & \left .
    \begin{aligned}
      & -(z'_{l,c,m,n})_a \leq 0 \\
      & (\Bar{z}_{l,c,m,n})_a -(z'_{l,c,m,n})_a  \leq 0 \\
     &  [(\Bar{u}_{l,c,m,n})_a - (\Bar{l}_{l,c,m,n})_a](z'_{l,c,m,n})_a -  (\Bar{u}_{l,c,m,n})_a(\Bar{z}_{l,c,m,n})_a + (\Bar{u}_{l,c,m,n})_a(\Bar{l}_{l,c,m,n})_a \leq 0 \\
    \end{aligned}
    \right \} &&, ~(c,m,n,a) \in \Bar{\mathcal{I}}_l \\
    \end{aligned}
    \right \} \\
& (z^M_{l,c,m,n})_{a+1} - [(z'_{l,c,m,n})_a + (z^M_{l,c,m,n})_a] = 0, \ (c,m,n) \in S_l^{IN}, \ a \in \llbracket 0, k_l^Mk_l^M-1 \rrbracket\\
& z_{l,c,m,n} - (z^M_{l,c,m,n})_{k_l^Mk_l^M} = 0, \ (c,m,n) \in S_l^{IN}\\
\end{aligned}
\right . \\
& \quad  \text{, where}  \ l \in \llbracket 2,k \rrbracket \\
& \Tilde{z}_{k,a} - \sum\limits_{(c,m,n) \in S_k^{IN}} (\hat{W}_{c,m,n})_a z_{k,c,m,n} = 0 , \ ~a \in \llbracket 0 ,a_k-1 \rrbracket \\
& \hat{\mathbf{z}}'_{l+1} - (\mathbf{W}_l\Tilde{\mathbf{z}}_l + \mathbf{b}_l) = \mathbf{0} ,\ ~l\in \llbracket k,k+t-1 \rrbracket \\
& \left \{
\begin{aligned}
&-\Tilde{z}_{l,a} = 0 && , a \in \hat{\mathcal{I}}_l^- \\
& \hat{z}'_{l,a} - \Tilde{z}_{l,a} = 0 && , a \in \hat{\mathcal{I}}_l^+ \\
&\left .
\begin{aligned}
& -\Tilde{z}_{l,a} \leq 0 \\
& \hat{z}'_{l,a} - \Tilde{z}_{l,a} \leq 0 \\
    	                                &(u'_{l,a} - l'_{l,a}) \Tilde{z}_{l,a} - u'_{l,a}\hat{z}'_{l,a} + u'_{l,a}l'_{l,a} \leq 0 \\
    	                           \end{aligned}
	                           \right \} && , a \in \hat{\mathcal{I}}_l \\
	                        \end{aligned}
	                  \right \}, \ ~ l \in \llbracket k+1,k+t-1 \rrbracket \\ \\
	       \textrm{variables} 
	                          \quad & {z}_{l,c,m,n}, \ (c,m,n)\in S_l, \ l \in \llbracket 1,k-1 \rrbracket \\
	                          \quad & {z}_{k,c,m,n},  \ (c,m,n) \in S_k^{IN}\\
	                          \quad & {\hat{z}}_{l,c,m,n}, \ (c,m,n) \in Q_l^{IN}, \ l \in  \llbracket 2,k \rrbracket\\
	                          \quad & {z}^R_{l,c,m,n}, \ (c,m,n) \in Q_l^{IN}, \ l \in  \llbracket 2,k \rrbracket\\
	                          \quad & ({z}^M_{l,c,m,n})_a, \ (c,m,n,a) \in S_l^{IN} \times \llbracket 0, k_l^Mk_l^M \rrbracket \ , l \in  \llbracket 2,k \rrbracket \\
	                          \quad & (\Bar{z}_{l,c,m,n})_a, \ (c,m,n,a) \in S_l^{IN} \times \llbracket 0, k_l^Mk_l^M-1 \rrbracket \ , l \in  \llbracket 2,k \rrbracket \\
	                          \quad & ({z}'_{l,c,m,n})_a, \ (c,m,n,a) \in S_l^{IN} \times \llbracket 0, k_l^Mk_l^M-1 \rrbracket \ , l \in  \llbracket 2,k \rrbracket \\
	                          \quad & \Tilde{\mathbf{z}}_l \in \mathbb{R}^{a_l}\ , l \in \llbracket k,k+t-1 \rrbracket \\
	                          \quad & \hat{\mathbf{z}}'_l \in \mathbb{R}^{a_l}\ , l \in  \llbracket k+1,k+t \rrbracket  \stepcounter{equation}\tag{\theequation}\label{primal}\\
\end{align*}
\end{figure*}

\subsubsection{Dual problem}\label{subsubsec:Dual Problem}
We associated the dual variables \citep{boyd2004convex} with each of the constraints as follows:
\begin{itemize}
  \item Norm-bound of the perturbation
        \begin{flalign*}
            &z_{1,c,m,n} - (x_{c,m,n} + \epsilon_{c,m,n}) \leq 0 \Rightarrow \xi^+_{c,m,n} \in \mathbb{R}_{\geq 0},   \ (c,m,n) \in S_1^{IN}\\
            &-z_{1,c,m,n} + (x_{c,m,n} - \epsilon_{c,m,n}) \leq 0  \Rightarrow \xi^-_{c,m,n}  \in \mathbb{R}_{\geq 0}, \ (c,m,n) \in S_1^{IN}
        \end{flalign*}
  \item Padding
        \begin{align*}
             &z_{l,c,m,n} = 0  \Rightarrow \nu^p_{l,c,m,n} \in \mathbb{R}, \ (c,m,n) \in S_l^{UD} \cup S_l^{LR}, \ {l\in \llbracket1,k-1\rrbracket} 
        \end{align*}
  \item Convolution 
        \begin{equation}\label{dual variable: conv}
            \begin{aligned}
             &\hat{z}_{l,c,m',n'} - b_{l-1,c} 
             -\sum\limits_{(\hat{c},m,n) \in S_{l-1}} h_{l-1,c}({\hat{c},m+p_{l-1}^{cv}-s_{l-1}^{cv}m',n+p_{l-1}^{cv}-s_{l-1}^{cv}n'})z_{l-1,\hat{c},m,n} =0 \\
             &\Rightarrow \nu_{{l},c,m',n'} \in  \mathbb{R}
             ,(c,m',n') \in F_{l}^{IN}, \ { l\in \llbracket2,k\rrbracket} \\
             \end{aligned}
        \end{equation}
  \item ReLU (In convolution part)
        \begin{equation}\label{dual variable: ReLU in CNN}
            \begin{aligned}
            &\left \{
                \begin{aligned}
                    &-z^R_{l,c,m,n} = 0 \Rightarrow \mu^R_{l,c,m,n} \in \mathbb{R} &&, \ (c,m,n) \in \mathcal{I}_l^- \\
                    &\hat{z}_{l,c,m,n} - z^R_{l,c,m,n} = 0 \Rightarrow \tau^R_{l,c,m,n} \in \mathbb{R} &&,\ (c,m,n) \in \mathcal{I}_l^+\\
                    &\left .
                        \begin{aligned}
                            &-z^R_{l,c,m,n} \leq 0 \Rightarrow \mu^R_{l,c,m,n} \in \mathbb{R}_{\geq 0} \\
                            &\hat{z}_{l,c,m,n} - z^R_{l,c,m,n} \leq 0 \Rightarrow \tau^R_{l,c,m,n} \in \mathbb{R}_{\geq 0} \\
                            & (\hat{u}_{l,c,m,n} - \hat{l}_{l,c,m,n})z^R_{l,c,m,n} - \hat{u}_{l,c,m,n}\hat{z}_{l,c,m,n}  +\hat{u}_{l,c,m,n}\hat{l}_{l,c,m,n} \leq 0 \Rightarrow \lambda^R_{l,c,m,n} \in \mathbb{R}_{\geq 0} \\
                        \end{aligned}
                    \right \}&&, \ (c,m,n) \in \mathcal{I}_l \\
                \end{aligned} 
            \right \} 
            ,\  { l \in \llbracket 2,k \rrbracket} \\
            \end{aligned}
        \end{equation}
  \item Maxpool 
  \newline
        \begin{equation}\label{dual variable: maxpool}
        \begin{aligned}
            &(z^M_{l,c,m,n})_0 = 0 \Rightarrow (\nu^M_{l,c,m,n})_0 \in \mathbb{R} ,\ (c,m,n) \in S_l^{IN}, \ l\in \llbracket 2,k \rrbracket \\
            &(\Bar{z}_{l,c,m,n})_{ik_l^M+j} - [ z^R_{l,c,s_l^Mm+i,s_l^Mn+j} - (z^M_{l,c,m,n})_{ik_l^M+j} ] = 0 \Rightarrow (\kappa_{l,c,m,n})_{ik_l^M+j} \in \mathbb{R}, \
            \left \{
                \begin{aligned}
                    &l\in \llbracket 2,k \rrbracket, \\
                    &(c,m,n) \in S_l^{IN},\\
                    &i \in \llbracket 0,k_l^M-1 \rrbracket,\\
                    &j \in \llbracket 0,k_l^M-1 \rrbracket\\
                \end{aligned}
            \right .\\
            &\left \{
                \begin{aligned}
                    &-(z'_{l,c,m,n})_a = 0 \Rightarrow (\mu^M_{l,c,m,n})_a \in \mathbb{R}  &&, \ (c,m,n,a) \in \bar{\mathcal{I}}_l^-\\
                    &(\Bar{z}_{l,c,m,n})_a - (z'_{l,c,m,n})_a = 0 \Rightarrow (\tau^M_{l,c,m,n})_a \in \mathbb{R} &&, \ (c,m,n,a) \in \bar{\mathcal{I}}_l^+\\
                    &\left .
                        \begin{aligned}
                            &-(z'_{l,c,m,n})_a \leq 0 \Rightarrow (\mu^M_{l,c,m,n})_a \in \mathbb{R}_{\geq 0} \\
                            &(\Bar{z}_{l,c,m,n})_a - (z'_{l,c,m,n})_a \leq 0 \Rightarrow (\tau^M_{l,c,m,n})_a \in \mathbb{R}_{\geq 0} \\
                            &[(\Bar{u}_{l,c,m,n})_a - (\Bar{l}_{l,c,m,n})_a](z'_{l,c,m,n})_a -  (\Bar{u}_{l,c,m,n})_a(\Bar{z}_{l,c,m,n})_a  +(\Bar{u}_{l,c,m,n})_a(\Bar{l}_{l,c,m,n})_a \\
                            &\quad \leq 0 \Rightarrow (\lambda^M_{l,c,m,n})_a \in \mathbb{R}_{\geq 0} \\
                        \end{aligned}
                    \right \} &&, \ (c,m,n,a) \in \bar{\mathcal{I}}_l \\
                \end{aligned}
            \right \} \\
            & \quad \text{, where }l\in \llbracket 2,k \rrbracket\\
            &(z^M_{l,c,m,n})_{a+1} -[ (z'_{l,c,m,n})_a + (z^M_{l,c,m,n})_a ] = 0\Rightarrow (\rho_{l,c,m,n})_{a+1} \in \mathbb{R}, \  (c,m,n,a) \in S_l^{IN}\times \llbracket 0, k_l^M k_l^M-1 \rrbracket, \ l\in \llbracket 2,k\rrbracket\\
            &z_{l,c,m,n} - (z^M_{l,c,m,n})_{k_l^Mk_l^M} = 0 \Rightarrow \beta_{l,c,m,n} \in \mathbb{R}, \  (c,m,n) \in S_l^{IN}, \ l\in \llbracket 2,k \rrbracket \\
        \end{aligned}
        \end{equation}
  \item Flatten
        \begin{flalign*}
            &\Tilde{z}_{k,a} - \sum\limits_{(c,m,n) \in S_k^{IN}} (\hat{W}_{c,m,n})_a z_{k,c,m,n} = 0 \Rightarrow \gamma_{k,a} \in \mathbb{R}, \  a \in \llbracket 0,a_k-1\rrbracket
        \end{flalign*}
  \item Fully connected
        \begin{flalign*}
            &\hat{\mathbf{z}}'_{l+1} - (\mathbf{W}_l\Tilde{\mathbf{z}}_l + \mathbf{b}_l)=0 \Rightarrow \bm{\nu}'_{{l+1}} \in \mathbb{R}^{a_{l+1}}, \  l\in \llbracket k,k+t-1 \rrbracket
        \end{flalign*}
  \item ReLU (In linear part) 
        \begin{equation}\label{dual variable - linear ReLu part}
            \left \{
                \begin{aligned}
                    &-\Tilde{z}_{l,a} = 0  \Rightarrow \mu'_{l,a} \in \mathbb{R} &&, a \in \hat{\mathcal{I}}_l^-\\
                    &\hat{z}'_{l,a} - \Tilde{z}_{l,a} = 0 \Rightarrow \tau'_{l,a} \in \mathbb{R} &&, a \in \hat{\mathcal{I}}_l^+\\
                    &\left .
                        \begin{aligned}
                            &-\Tilde{z}_{l,a} \leq 0  \Rightarrow \mu'_{l,a} \in \mathbb{R}_{\geq 0}  \\
                            &\hat{z}'_{l,a} - \Tilde{z}_{l,a} \leq 0 \Rightarrow \tau'_{l,a} \in \mathbb{R}_{\geq 0} \\
                            &(u'_{l,a} - l'_{l,a}) \Tilde{z}_{l,a} - u'_{l,a}\hat{z}'_{l,a} +u'_{l,a}l'_{l,a} \leq 0\Rightarrow \lambda'_{l,a} \in \mathbb{R}_{\geq 0} \\
                        \end{aligned}
                    \right \} &&, a \in \hat{\mathcal{I}}_l
                \end{aligned}
            \right \} , \ l \in \llbracket k+1,k+t-1 \rrbracket  \\
        \end{equation}
\end{itemize}
Let $\Theta$ denote the collection of dual variables, the dual problem of can be derived as \eqref{dual} {with dual variables \eqref{dual variables}.}\\
\begin{subequations}\label{dual}
    \begin{align}
        &\text{maximize} \nonumber
        \\
        \begin{split}
            \quad g({\Theta}) 
             &= \sum\limits_{(c,m,n) \in S_1} -x_{c,m,n}(\xi_{c,m,n}^+ - \xi_{c,m,n}^-) - \sum\limits_{(c,m,n) \in S_1} \epsilon_{c,m,n}(\xi_{c,m,n}^+ + \xi_{c,m,n}^-) \\
               &+\sum\limits_{\substack{l \in \llbracket 2,k \rrbracket \\(c,m,n)\in Q_l^{IN}}} \lambda^R_{l,c,m,n}(\hat{u}_{l,c,m,n}\hat{l}_{l,c,m,n}) 
             + \sum\limits_{\substack{l \in \llbracket 2,k \rrbracket\\ (c,m,n) \in S_l^{IN} \\ a \in \llbracket 0 , k_l^M k_l^M-1 \rrbracket}}(\lambda^M_{l,c,m,n})_a[(\Bar{u}_{l,c,m,n})_a(\Bar{l}_{l,c,m,n})_a] \\
               &+ \sum\limits_{l\in \llbracket k+1 ,k+t-1\rrbracket} {\bm{\lambda}'_l}^T[\bm{u}'_l \odot \bm{l}'_l]- \sum\limits_{\substack{l\in \llbracket 1 , k-1\rrbracket \\ (c,m,n)\in                                    Q_{l+1}^{IN} }} \nu_{l+1,c,m,n}b_{l,c} -\sum\limits_{l\in  \llbracket k , k+t-1 \rrbracket} {\bm{\nu}'_{l+1}}^T \bm{b}_l  \\ \\
        \end{split}\label{eq:dual obj} \\ 
        &\text{subject to} \nonumber
        \\
        \begin{split}
              \quad &\bm{\nu}'_{k+t} = \bm{d} \\
        \end{split}\label{eq:dual con1} \\
        \begin{split}
             \quad &\xi_{c,m,n}^+ - \xi_{c,m,n}^- + \nu_{1,c,m,n}^P = \sum\limits_{(\hat{c},m',n') \in Q_2^{IN}} \nu_{2,\hat{c},m',n'}h_{1,\hat{c}}({c,m+p_1^{cv}-s_1^{cv}m',n+p_1^{cv}-s_1^{cv}n'}), \ (c,m,n) \in S_1 \\
        \end{split}\label{eq:dual con2} \\
        \begin{split}
             \quad & 
             \nu_{l,c,m,n}^P + \beta_{l,c,m,n} = 
              \sum\limits_{(\hat{c},m',n')\in Q_{l+1}^{IN}}  \nu_{l+1,\hat{c},m',n'}h_{l,\hat{c}}({c,m+p_l^{cv}-s_l^{cv}m',n+p_l^{cv}-s_l^{cv}n'}),  
                     \begin{cases}
                          {l \in \llbracket 2,k-1 \rrbracket}, \\
                          (c,m,n) \in S_l \\
                     \end{cases} \\
        \end{split}\label{eq:dual con3} \\
        \begin{split}
             \quad & \beta_{k,c,m,n} = \sum\limits_{a=0}^{a_k-1} (\hat{W}_{c,m,n})_a \gamma_{k,a}, \ (c,m,n) \in S_k^{IN} \\
        \end{split}\label{eq:dual con4}  \\
        \begin{split}
             \quad &\nu_{l,c,m,n} = -\tau^R_{l,c,m,n} + \lambda^R_{l,c,m,n}\hat{u}_{l,c,m,n} , \ \left \{
                            \begin{aligned}
                                & {l\in \llbracket 2,k \rrbracket}, \\
                                & (c,m,n)\in Q_l^{IN} \\
                            \end{aligned}
                        \right . \\
        \end{split}\label{eq:dual con5} \\
        \begin{split}
              \quad & - \mu^R_{l,c,m,n} -\tau^R_{l,c,m,n} +\lambda^R_{l,c,m,n}(\hat{u}_{l,c,m,n}-\hat{l}_{l,c,m,n}) = \sum\limits_{\substack{(i,j) \in \llbracket 0, k_l^M-1 \rrbracket \times \llbracket 0, k_l^M-1 \rrbracket \\ (c,\frac{m-i}{s_l^M},\frac{n-j}{s_l^M}) \in S_l^{IN}}} (\kappa_{l,c,\frac{m-i}{s_l^M},\frac{n-j}{s_l^M}})_{i k_l^M+j}, \ 
                        \left \{
                            \begin{aligned}
                                & {l\in \llbracket 2,k \rrbracket}, \\
                                & (c,m,n)\in Q_l^{IN} \\
                            \end{aligned}
                        \right . \\
        \end{split}\label{eq:dual con6} \\
        \begin{split}
             \quad &(\nu_{l,c,m,n}^M)_0 = -(\kappa_{l,c,m,n})_0 + (\rho_{l,c,m,n})_1 , \
            \left\{
            \begin{aligned}
                & { l \in \llbracket 2,k \rrbracket}, \\
                &(c,m,n) \in S_l^{IN}
            \end{aligned}
            \right. \\
        \end{split}\label{eq:dual con7} \\
        \begin{split}
             \quad &(\rho_{l,c,m,n})_{k_l^M k_l^M} = \beta_{l,c,m,n}, \ 
             \left\{
            \begin{aligned}
                & { l \in \llbracket 2,k \rrbracket}, \\
                &(c,m,n) \in S_l^{IN}
            \end{aligned}
            \right. \\
        \end{split}\label{eq:dual con8} \\
        \begin{split}
             \quad & (\rho_{l,c,m,n})_a = (\rho_{l,c,m,n})_{a+1} - (\kappa_{l,c,m,n})_a, \
             \left\{
            \begin{aligned}
                & { l \in \llbracket 2,k \rrbracket}, \\
                &(c,m,n) \in S_l^{IN}, \\
                & { a \in \llbracket 1, k_l^M k_l^M-1 \rrbracket}
            \end{aligned}
            \right. \\
        \end{split}\label{eq:dual con9} \\
        \begin{split}
             \quad & (\kappa_{l,c,m,n})_a = -(\tau_{l,c,m,n}^M)_a + (\lambda_{l,c,m,n}^M)_a (\Bar{u}_{l,c,m,n})_a, \
            \left\{
            \begin{aligned}
                & { l \in \llbracket 2,k \rrbracket}, \\
                &(c,m,n) \in S_l^{IN}, \\
                & { a \in \llbracket 0, k_l^M k_l^M-1 \rrbracket}
            \end{aligned}
            \right. \\
        \end{split}\label{eq:dual con10} \\
        \begin{split}
             \quad & (\rho_{l,c,m,n})_{a+1} = -[(\mu_{l,c,m,n}^M)_a + (\tau_{l,c,m,n}^M)_a] + (\lambda_{l,c,m,n}^M)_a[(\Bar{u}_{l,c,m,n})_a - (\Bar{l}_{l,c,m,n})_a], \
                        \left\{
                        \begin{aligned}
                            & { l \in \llbracket 2,k \rrbracket}, \\
                            &(c,m,n) \in S_l^{IN}, \\
                            & { a \in \llbracket 0, k_l^M k_l^M-1 \rrbracket}
                        \end{aligned}
                        \right. \\
        \end{split}\label{eq:dual con11} \\
        \begin{split}
             \quad & \mathbf{W}^T_l \bm{\nu}'_{l+1} = -(\bm{\mu}'_l+\bm{\tau}'_l) + \bm{\lambda}'_l \odot (\bm{u}'_l-\bm{l}'_l), \ { l \in  \llbracket k+1,k+t-1 \rrbracket} \\
        \end{split}\label{eq:dual con12} \\
        \begin{split}
             \quad & \bm{\nu}'_l = -\bm{\tau}'_l + \bm{\lambda}'_l \odot \bm{u}'_l, \ { l\in \llbracket k+1,k+t-1 \rrbracket} \\
        \end{split}\label{eq:dual con13} \\
        \begin{split}
             \quad & \bm{\gamma}_k = \mathbf{W}_k^T \bm{\nu}'_{k+1} \\
        \end{split}\label{eq:dual con14} \\
    \end{align}
\end{subequations}
\begin{subequations}\label{dual variables}
\begin{align}
    \begin{split}
        \quad & \left \{
                    \begin{aligned}
                        & \xi^+_{c,m,n}, \xi^-_{c,m,n} \in \mathbb{R}_{\geq 0}, \ (c,m,n) \in S_1^{IN} \\ 
                        & \xi^+_{c,m,n}, \xi^-_{c,m,n} = 0, \ (c,m,n) \in S_1^{UD} \cup S_1^{LR} \\
                    \end{aligned}
                \right . \\
    \end{split}\label{eq:dual var1} \\
    \begin{split}
        \quad & \left \{
                    \begin{aligned}
                        & \nu^P_{l,c,m,n} = 0 , \ (c,m,n) \in S_l^{IN} \\ 
                        & \nu^P_{l,c,m,n} \in \mathbb{R}, \ (c,m,n) \in S_l^{UD} \cup S_l^{LR} \\
                    \end{aligned}
                \right \} , \ { l\in \llbracket 1,k-1 \rrbracket} \\
    \end{split}\label{eq:dual var2} \\
    \begin{split}
        \quad & \nu_{l,c,m,n} \in \mathbb{R}, \ (c,m,n) \in Q_l^{IN},         \ { l\in \llbracket 2,k \rrbracket} \\
    \end{split}\label{eq:dual var nu} \\
    \begin{split}
        \quad & \left \{
                    \begin{aligned}
                        & \mu^R_{l,c,m,n}  \in \mathbb{R}, \ \tau^R_{l,c,m,n} = 0 , \ \lambda^R_{l,c,m,n} = 0
                        &&, (c,m,n) \in \mathcal{I}_l^- \\
                        & \mu^R_{l,c,m,n}  \geq 0, \ \tau^R_{l,c,m,n} \geq 0 , \ \lambda^R_{l,c,m,n} \geq 0
                        &&, (c,m,n) \in \mathcal{I}_l \\
                        & \mu^R_{l,c,m,n} = 0, \ \tau^R_{l,c,m,n} \in \mathbb{R} , \ \lambda^R_{l,c,m,n} = 0
                        &&, (c,m,n) \in \mathcal{I}_l^+\\
                    \end{aligned}
                \right . , \ { l\in \llbracket 2,k \rrbracket} \\
    \end{split}\label{eq:dual var3} \\
    \begin{split}
        \quad & (\nu^M_{l,c,m,n})_0 \in \mathbb{R}, \ (c,m,n) \in S_l^{IN},         \ { l\in \llbracket 2,k \rrbracket} \\
    \end{split}\label{eq:dual var4} \\
    \begin{split}
        \quad & (\kappa_{l,c,m,n})_a \in \mathbb{R}, \ (c,m,n,a) \in S_l^{IN} \times \llbracket 0,k_l^M k_l^M-1 \rrbracket, \ { l\in \llbracket 2,k \rrbracket} \\
    \end{split}\label{eq:dual var5} \\
    \begin{split}
        \quad & \left \{
                    \begin{aligned}
                        & (\mu^M_{l,c,m,n})_a  \in \mathbb{R}, \ (\tau^M_{l,c,m,n})_a = 0 , \ (\lambda^M_{l,c,m,n})_a = 0
                        &&, (c,m,n,a) \in \bar{\mathcal{I}}_l^- \\
                        & (\mu^M_{l,c,m,n})_a  \geq 0, \ (\tau^M_{l,c,m,n})_a \geq 0 , \ (\lambda^M_{l,c,m,n})_a \geq 0
                        &&, (c,m,n,a) \in \bar{\mathcal{I}}_l \\
                        & (\mu^M_{l,c,m,n})_a = 0, \ (\tau^M_{l,c,m,n})_a \in \mathbb{R} , \ (\lambda^M_{l,c,m,n})_a = 0
                        &&, (c,m,n,a) \in \bar{\mathcal{I}}_l^+\\
                    \end{aligned}
                \right \} \ , { l\in \llbracket 2,k \rrbracket} \\
    \end{split}\label{eq:dual var6} \\
    \begin{split}
        \quad & (\rho_{l,c,m,n})_a \in \mathbb{R}, \ (c,m,n,a) \in S_l^{IN} \times \llbracket 1,k_l^M k_l^M \rrbracket, \ { l\in \llbracket 2,k \rrbracket}\\
    \end{split}\label{eq:dual var7} \\
    \begin{split}
        \quad & \left \{
                    \begin{aligned}
                        & \beta_{l,c,m,n} \in \mathbb{R}, \  (c,m,n) \in S_l^{IN} \\
                        & \beta_{l,c,m,n} = 0,  \ (c,m,n) \in S_l^{UD}\cup S_l^{LR} \\
                    \end{aligned} 
                \right \}, \ { l\in \llbracket 2,k \rrbracket} \\
    \end{split}\label{eq:dual var8} \\
    \begin{split}
        \quad & \bm{\gamma}_k \in \mathbb{R}^{a_k} \\
    \end{split}\label{eq:dual var9} \\
    \begin{split}
        \quad & \bm{\nu}'_l \in \mathbb{R}^{a_l}, \  { l\in \llbracket k+1,k+t \rrbracket} \\
    \end{split}\label{eq:dual var10} \\
    \begin{split}
        \quad & \left \{
                    \begin{aligned}
                        & \mu'_{l,a}  \in \mathbb{R}, \ \tau'_{l,a} = 0 , \ \lambda'_{l,a} = 0
                        &&, a \in \hat{\mathcal{I}}_l^- \\
                        & \mu'_{l,a}  \geq 0, \ \tau'_{l,a} \geq 0 , \ \lambda'_{l,a} \geq 0
                        &&, a \in \hat{\mathcal{I}}_l \\
                        & \mu'_{l,a} = 0, \ \tau'_{l,a} \in \mathbb{R} , \ \lambda'_{l,a} = 0
                        &&, a \in \hat{\mathcal{I}}_l^+\\
                    \end{aligned}
                \right \} , \ { l\in \llbracket k+1,k+t-1 \rrbracket} \\
    \end{split}\label{eq:dual var11}
\end{align}
\end{subequations}

\subsubsection{CNN dual network}\label{subsubsec:CNN Dual Network}
\begin{figure}[ht]
    \centering
    \includegraphics[width = 0.5 \textwidth]{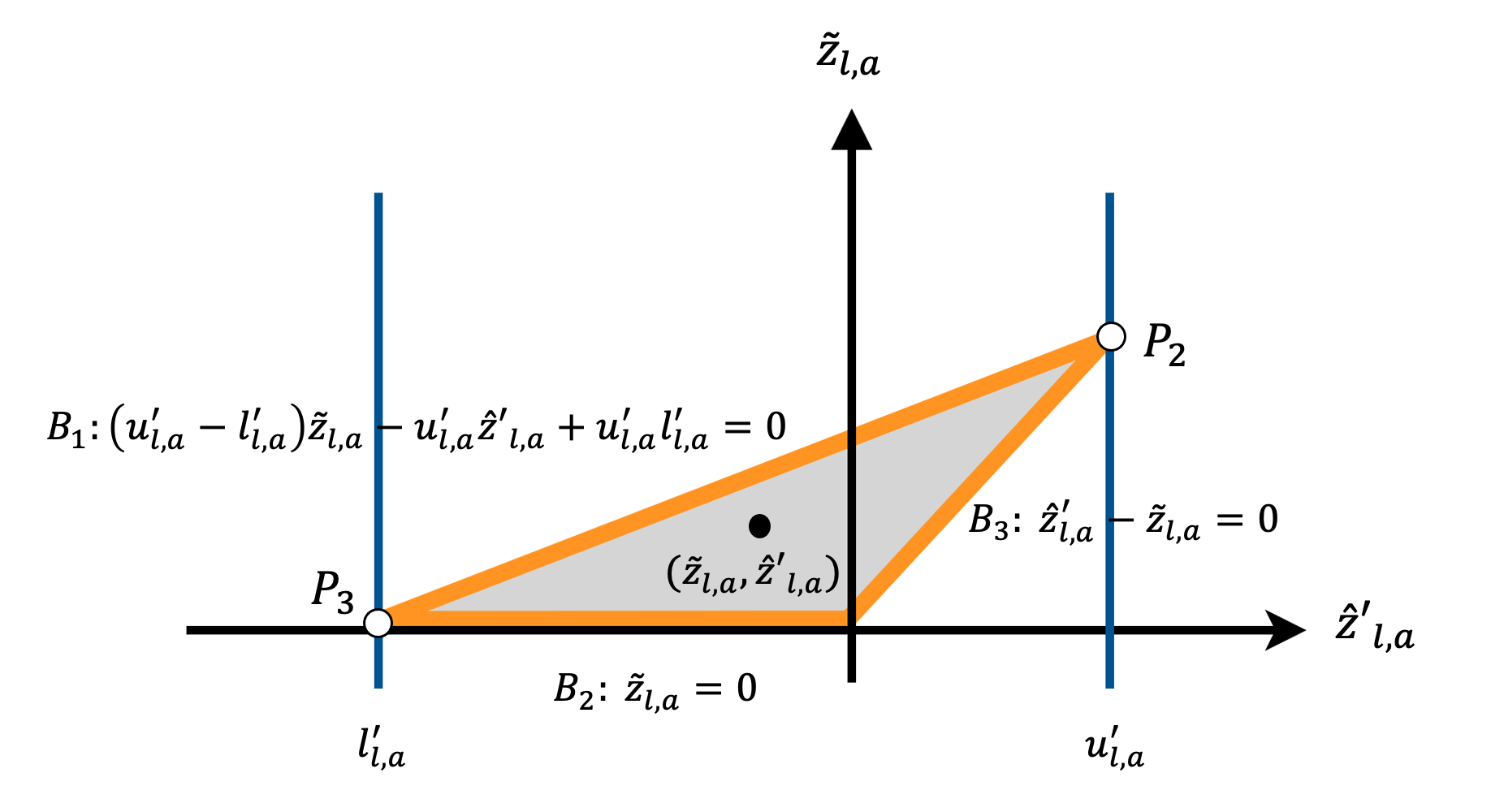}
    \caption{
    {
        Recall \eqref{upper-lower of hat_z'} that the primal feasible solution for $(\hat{z}'_{l,a} , \Tilde{z}_{l,a})$ must lie within the grayed area (not including points $P_2$ and $P_3$). If $(\hat{z}'_{l,a} , \Tilde{z}_{l,a})$ does not lie on boundary $B_1$, then \eqref{AAA-c} implies $\lambda'_{l,a}=0$. If $(\hat{z}'_{l,a}, \Tilde{z}_{l,a})$ lies on $B_1$, then \eqref{AAA-a} and \eqref{AAA-b} imply $\mu'_{l,a} = \tau'_{l,a}=0$.
    }}
    \label{fig:convex_relaxation_fgh.png}
\end{figure}

In this section, we introduce how to successively solve the optimal dual variables in a way similar to a feed-forward CNN, which is referred as the dual network. We start from directly solving $\nu'_{k+t} = d$ (cf. \eqref{eq:dual con1}). By substituting \eqref{eq:dual var11} into \eqref{eq:dual con12} and \eqref{eq:dual con13}, we can easily express $\mu'_{l,a}, \tau'_{l,a}, \lambda'_{l,a}$, { $\nu'_{l,a}$} in terms of $W_l^T\nu'_{l+1}$ when $a$ either belongs to $\hat{\mathcal{I}}^-_l$ or $\hat{\mathcal{I}}^+_l$, namely
\begin{equation*}
    \begin{aligned}
        &\mu'_{l,a} = -[W_l^T\nu'_{l+1}]_a, \ \tau'_{l,a}=0, \ \lambda'_{l,a}=0, \ \nu'_{l,a}=0 && , \ a \in \hat{\mathcal{I}}^-_l \\
        &\mu'_{l,a} = 0, \ \tau'_{l,a} = -[W_l^T\nu'_{l+1}]_a, \ \lambda'_{l,a}=0 , \ \nu'_{l,a} = [W_l^T\nu'_{l+1}]_a && , \ a \in \hat{\mathcal{I}}^+_l.
    \end{aligned}
\end{equation*}
For $a \in \hat{\mathcal{I}_l}$, impose complementary slackness constraints to \eqref{dual variable - linear ReLu part} yields
\begin{subequations}
        \begin{align}
            \begin{split}
                &\mu'_{l,a}(-\Tilde{z}_{l,a})=0
            \end{split}\label{AAA-a}\\
            \begin{split}
                &\tau'_{l,a}(\hat{z}'_{l,a} - \Tilde{z}_{l,a}) =0
            \end{split}\label{AAA-b}\\
            \begin{split}
                &\lambda'_{l,a}[ (u'_{l,a}-l'_{l,a})\Tilde{z}_{l,a} -u'_{l,a}\hat{z}'_{l,a} + u'_{l,a}l'_{l,a}]=0
            \end{split}\label{AAA-c}
        \end{align}
\end{subequations}
Note that as illustrated in Fig.~\ref{fig:convex_relaxation_fgh.png}, either $\lambda'_{l,a}=0$ or { $\mu'_{l,a} = \tau'_{l,a} = 0$}. Recall that $\mu'_{l,a}, \tau'_{l,a}, \lambda'_{l,a}$ are non-negative (cf. \eqref{eq:dual var11}). If $\lambda'_{l,a}=0$, then by \eqref{eq:dual con12},
\begin{equation*}
    \begin{aligned}
        &\mu'_{l,a}+\tau'_{l,a} = -(W_l^T\nu'_{l+1})_a = [(W_l^T\nu'_{l+1})_a]_- \\
        &\lambda'_{l,a}(u'_{l,a}-l'_{l,a}) = 0 = [(W_l^T\nu'_{l+1})_a]_+
    \end{aligned}
\end{equation*}
where we use notation $[\xi]_+ = \xi \vee 0$ and $[\xi]_- = (-\xi) \vee 0$. If $\mu'_{l,a} = \tau'_{l,a} = 0$, then by \eqref{eq:dual con12}
\begin{equation*}
    \begin{aligned}
        & \lambda'_{l,a}(u'_{l,a}-l'_{l,a}) = (W_l^T\nu'_{l+1})_a = [(W_l^T\nu'_{l+1})_a]_+ \\
        & \mu'_{l,a} + \tau'_{l,a} = 0 = [(W_l^T\nu'_{l+1})_a]_-
    \end{aligned} \ .
\end{equation*}
As such, for { $a \in \hat{\mathcal{I}}_l$}, we can write
\begin{equation}\label{eq:simplified result by using complementary slackness}
    \begin{aligned}
        (\mu'_l + \tau'_l)_a &= [(W^T_l\nu'_{l+1})_a]_- \\
        (\lambda'_l \odot (u'_l-l'_l))_a &= [(W^T_l\nu'_{l+1})_a]_+ 
    \end{aligned} \ .
\end{equation}
Hence, by introducing additional variables $\alpha'_{l,a} \in [0,1]$, we can rewrite \eqref{eq:simplified result by using complementary slackness} and \eqref{eq:dual con13} as
\begin{equation}\label{eq:mu' tau' nu' gotten from complementary slackness}
    \begin{aligned}
        \tau'_{l,a} &=  \alpha'_{l,a}[(W^T_l\nu'_{l+1})_a]_-\\
        \mu'_{l,a} &= (1-\alpha'_{l,a})[(W^T_l\nu'_{l+1})_a]_-\\ 
        \lambda'_{l,a} &= \frac{1}{u'_{l,a}-l'_{l,a}}[(W^T_l\nu'_{l+1})_a]_+ \\
        \nu'_{l,a} &= \frac{u'_{l,a}}{u'_{l,a}-l'_{l,a}}[(W^T_l\nu'_{l+1})_a]_+ - \alpha'_{l,a}[(W^T_l\nu'_{l+1})_a]_-  \ .
    \end{aligned}
\end{equation}
Thus, with $\nu'_{l+1}$ at hand { and introduce variable $\hat{\nu}'_{l,a} = (W^T_l\nu'_{l+1})_a$}, we can solve $\nu'_l$ as follows:
\begin{equation}
    \begin{aligned}
        \mu'_{l,a} &= \left \{
            \begin{aligned}
                &-\hat{\nu}'_{l,a}                     &&, a \in \hat{\mathcal{I}}_l^- \\
                &0                                     &&, a \in \hat{\mathcal{I}}_l^+ \\
                &(1-\alpha'_{l,a})[\hat{\nu}'_{l,a}]_- \ \ \ \ \ \ \ &&, a \in \hat{\mathcal{I}}_l \\
            \end{aligned} 
        \right . \\
        \tau'_{l,a} &= \left \{
            \begin{aligned}
                &0                                     &&, a \in \hat{\mathcal{I}}_l^- \\
                &-\hat{\nu}'_{l,a}                     &&, a \in \hat{\mathcal{I}}_l^+ \\
                &\alpha'_{l,a}[\hat{\nu}'_{l,a}]_-  \ \ \ \ \ \ \ \ \  \ \ \ \ &&, a \in \hat{\mathcal{I}}_l \\
            \end{aligned} 
        \right . \\
        \lambda'_{l,a} &= \left \{
            \begin{aligned}
                &0                                     &&, a \in \hat{\mathcal{I}}_l^- \\
                &0                                     &&, a \in \hat{\mathcal{I}}_l^+ \\
                &\frac{1}{u'_{l,a}-l'_{l,a}}[\hat{\nu}'_{l,a}]_+  \ \ \ \ \ \     &&, a \in \hat{\mathcal{I}}_l \\
            \end{aligned} 
        \right . \\
    \end{aligned}
\end{equation}
\begin{equation} \label{eq:nu in diff set}
    \begin{aligned}
        &\nu'_{l,a} = \left \{
                        \begin{aligned}
                            & 0 &&, a \in \hat{\mathcal{I}}_l^- \\
                            & \hat{\nu}'_{l,a} &&, a \in \hat{\mathcal{I}}_l^+ \\ 
                            & \frac{u'_{l,a}}{u'_{l,a}-l'_{l,a}}[\hat{\nu}'_{l,a}]_+ - \alpha'_{l,a}[\hat{\nu}'_{l,a}]_- &&, a \in \hat{\mathcal{I}}_l \\
                        \end{aligned}
                     \right \} \\
        &, \ { l \in  \llbracket k+1,k+t-1 \rrbracket} \\
    \end{aligned}
\end{equation}
After obtaining $\bm{\nu}'_{k+1}$, we can use it to calculate $\bm{\gamma}_k$ by \eqref{eq:dual con14} and then use \eqref{eq:dual con4} to reshape $\bm{\gamma}_k$ as a three-dimensional map $\bm{\beta}_k$. The constraints \eqref{eq:dual con8} also indicates that the value of $(\rho_{l,c,m,n})_{k_l^M k_l^M}$ is equal to $\beta_{l,c,m,n}$. To sum up, with currently available information, we can compute
\begin{equation}\label{eq:get gamma}
    \gamma_k = W_k^T\bm{\nu}'_{k+1} \\
\end{equation}
\begin{equation}\label{eq:get beta from gamma}
    \beta_{k,c,m,n} = \sum\limits_{a=0}^{a_k-1} (\hat{W}_{c,m,n})_a \gamma_{k,a} , \ (c,m,n) \in S_k^{IN} \\
\end{equation}
\begin{equation}\label{eq: beta equal to rho}
    (\rho_{k,c,m,n})_{k_l^M k_l^M} = \beta_{k,c,m,n}, \ (c,m,n) \in S_l^{IN} .
\end{equation}
We then use $(\rho_{k,c,m,n})_{k_k^M k_k^M}$ to solve other dual variables. Observe that \eqref{eq:dual con10}, \eqref{eq:dual con11} and \eqref{eq:dual var nu} take similar form as \eqref{eq:dual con12}, \eqref{eq:dual con13} and \eqref{eq:dual var11}. Thus, by similar derivation, we may introduce additional variables $(\alpha^M_{l,c,m,n})_a \in [0,1]$ and obtain \eqref{get muM, tauM, lambdaM}, \eqref{eq: get kappa}.
\begin{equation}\label{get muM, tauM, lambdaM}
    {
    \begin{aligned}
        (\mu^M_{l,c,m,n})_a &= \left \{
            \begin{aligned}
                &-(\rho_{l,c,m,n})_{a+1}                              &&, (c,m,n,a) \in \bar{\mathcal{I}}_l^- \\
                &0                                                    &&, (c,m,n,a) \in \bar{\mathcal{I}}_l^+ \\
                &(1-(\alpha^M_{l,c,m,n})_a)[(\rho_{l,c,m,n})_{a+1}]_- \ \ \ \ \ \ \ \ \ &&, (c,m,n,a) \in \bar{\mathcal{I}}_l \\
            \end{aligned} 
        \right . \\
        (\tau^M_{l,c,m,n})_a &= \left \{
            \begin{aligned}
                &0                                                  &&, (c,m,n,a) \in \bar{\mathcal{I}}_l^- \\
                &-(\rho_{l,c,m,n})_{a+1}                            &&, (c,m,n,a) \in \bar{\mathcal{I}}_l^+ \\
                &(\alpha^M_{l,c,m,n})_a[(\rho_{l,c,m,n})_{a+1}]_-  \ \ \ \ \ \ \ \ \ \ \ \ \ \ \ \ \  &&, (c,m,n,a) \in \bar{\mathcal{I}}_l \\
            \end{aligned} 
        \right . \\
        (\lambda^M_{l,c,m,n})_a &= \left \{
            \begin{aligned}
                &0                                      &&, (c,m,n,a) \in \bar{\mathcal{I}}_l^- \\
                &0                                      &&, (c,m,n,a) \in \bar{\mathcal{I}}_l^+ \\
                &\frac{1}{{(\bar{u}_{l,c,m,n})_a} - (\bar{l}_{l,c,m,n})_a}[{(\rho_{l,c,m,n})_{a+1}}]_+ &&, (c,m,n,a) \in \bar{\mathcal{I}}_l \\
            \end{aligned} 
        \right . \\
    \end{aligned}
    }
\end{equation}
\begin{equation}\label{eq: get kappa}
    {
    \begin{aligned}
    (\kappa_{l,c,m,n})_a &= \left \{
                                \begin{aligned}
                                    & 0 &&, (c,m,n,a) \in \bar{\mathcal{I}}_l^- \\
                                    & (\rho_{l,c,m,n})_{a+1} &&, (c,m,n,a) \in \bar{\mathcal{I}}_l^+ \\ 
                                    & \frac{(\bar{u}_{l,c,m,n})_a}{(\bar{u}_{l,c,m,n})_a-(\bar{l}_{l,c,m,n})_a}[(\rho_{l,c,m,n})_{a+1}]_+ - (\alpha_{l,c,m,n}^M)_a[(\rho_{l,c,m,n})_{a+1}]_- &&, (c,m,n,a) \in \bar{\mathcal{I}}_l \\
                                \end{aligned}
                            \right \} \\
    \end{aligned}
    }
\end{equation}
Thus, $(\rho_{l,c,m,n})_a$ can be obtained by \eqref{eq:dual con9}, namely:
\begin{equation}\label{eq: relationship between kappa and Rho}
    (\rho_{l,c,m,n})_a = (\rho_{l,c,m,n})_{a+1} - (\kappa_{l,c,m,n})_a \  . 
\end{equation}
With $(\kappa_{l,c,m,n})_a$ at hand, we may solve other dual variables as elaborated below. We first define
\begin{equation} \label{eq: kaapa hat}
    \begin{aligned}
        &\hat{\kappa}_{l,c,m,n} = \sum\limits_{\substack{(i,j) \in \llbracket 0, k_l^M-1 \rrbracket \times \llbracket 0, k_l^M-1 \rrbracket \\ (c,\frac{m-i}{s_l^M},\frac{n-j}{s_l^M}) \in S_l^{IN}}} (\kappa_{l,c,\frac{m-i}{s_l^M},\frac{n-j}{s_l^M}})_{i k_l^M+j}, \\
        & l \in \llbracket 2,k \rrbracket, \
            (c,m,n)\in Q_l^{IN}  \ .
    \end{aligned}
\end{equation}
Observe that \eqref{eq:dual con5}, \eqref{eq:dual con6} and \eqref{eq:dual var3} take similar form as \eqref{eq:dual con12}, \eqref{eq:dual con13} and \eqref{eq:dual var11}. Thus by similar derivation, we may introduce additional variables $\alpha^R_{l,c,m,n} \in [0,1]$ { and obtain \eqref{eq: muR, tauR, lambdR} and \eqref{eq: nu conv} }
\begin{equation}\label{eq: muR, tauR, lambdR}
    \begin{aligned}
        \mu^R_{l,c,m,n} &= \left \{
            \begin{aligned}
                &-\hat{\kappa}_{l,c,m,n}                              &&, (c,m,n) \in \mathcal{I}_l^- \\
                &0                                                    &&, (c,m,n) \in \mathcal{I}_l^+ \\
                &{(1-\alpha^R_{l,c,m,n})[\hat{\kappa}_{l,c,m,n}]_-} \ &&, (c,m,n) \in \mathcal{I}_l \\
            \end{aligned} 
        \right . \\
        \tau^R_{l,c,m,n} &= \left \{
            \begin{aligned}
                &0                                                  &&, (c,m,n) \in \mathcal{I}_l^- \\
                &-\hat{\kappa}_{l,c,m,n}                            &&, (c,m,n) \in \mathcal{I}_l^+ \\
                &\alpha^R_{l,c,m,n}[\hat{\kappa}_{l,c,m,n}]_-   \ \ \ \ \ \ \ \ \ \ \ \ \  &&, (c,m,n) \in \mathcal{I}_l \\
            \end{aligned} 
        \right . \\
        \lambda^R_{l,c,m,n} &= \left \{
            \begin{aligned}
                &0                                      &&, (c,m,n) \in \mathcal{I}_l^- \\
                &0                                      &&, (c,m,n) \in \mathcal{I}_l^+ \\
                &\frac{1}{\hat{u}_{l,c,m,n}-\hat{l}_{l,c,m,n}}[\hat{\kappa}_{l,c,m,n}]_+ &&, (c,m,n) \in \mathcal{I}_l \\
            \end{aligned} 
        \right . \\
    \end{aligned}
\end{equation}
\begin{equation}\label{eq: nu conv}
{   
    \nu_{l,c,m,n}= \left \{
                            \begin{aligned}
                                & 0 &&, (c,m,n) \in \mathcal{I}_l^- \\
                                & \hat{\kappa}_{l,c,m,n} &&,  (c,m,n) \in \mathcal{I}_l^+ \\ 
                                & \frac{\hat{u}_{l,c,m,n}}{\hat{u}_{l,c,m,n} - \hat{l}_{l,c,m,n}}[\hat{\kappa}_{l,c,m,n}]_+ - \alpha_{l,c,m,n}^R[\hat{\kappa}_{l,c,m,n}]_- &&, (c,m,n) \in \mathcal{I}_l \\
                            \end{aligned}
                    \right \} \\
}
\end{equation}
With $\nu_{l+1,c,m,n}$ at hand, we may solve $\beta_{l,c,m,n}$ with \eqref{eq:dual con3} (Recall that $\nu^P_{l,c,m,n} = 0$ when $(c,m,n) \in S_l^{IN}$ by  \eqref{eq:dual con6})
\begin{equation}\label{eq: beta equal nu hat}
    \beta_{l,c,m,n} = \hat{\nu}_{l,c,m,n} , \ l \in \llbracket 2, k-1 \rrbracket, (c,m,n) \in S_l^{IN}
\end{equation}
%
%
\begin{equation}\label{eq:nu hat get from conv nu}
{\hat{\nu}_{l,c,m,n} 
= \sum\limits_{(\hat{c},m',n') \in Q_{l+1}^{IN}} \nu_{l+1,\hat{c},m',n'} h_{l,\hat{c}}({c,m+p_l^{cv}-s_l^{cv}m',n+p_l^{cv}-s_l^{cv}n'}), \ { l \in \llbracket 1, k-1 , \ (c,m,n) \in S_l^{IN}}
}
\end{equation}
where we define $\hat{\nu}_{l,c,m,n}$ in \eqref{eq:nu hat get from conv nu}. We thus obtain {$(\rho_{l-1,c,m,n})_{k_l^Mk_l^M}$} by \eqref{eq:dual con8}. To sum up, we may consecutively solve the dual variables associated with the convolution layers in a layer-by-layer manner:
\begin{equation}
\begin{aligned}
     &\beta_{l,c,m,n} \xrightarrow{\eqref{eq: beta equal to rho}} (\rho_{l,c,m,n})_{k_l^Mk_l^M} \xrightarrow{\eqref{eq: get kappa}, \eqref{eq: relationship between kappa and Rho}} (\kappa_{l,c,m,n})_a , \ (\rho_{l,c,m,n})_a  \\ 
     &\xrightarrow{\eqref{eq: kaapa hat}} \hat{\kappa}_{l,c,m,n} \xrightarrow{\eqref{eq: nu conv}} \nu_{l,c,m,n} \xrightarrow{\eqref{eq:nu hat get from conv nu}} \hat{\nu}_{l-1,c,m,n} \xrightarrow{\eqref{eq: beta equal nu hat}} \beta_{l-1,c,m,n}  \ .     
\end{aligned}    
\end{equation}
After obtaining $\nu_2$, recall \eqref{eq:dual con2} as well as constraints \eqref{eq:dual var1} and \eqref{eq:dual var2}, it follows that
%
\begin{equation}\label{eq: rewrite xi constraint}
    \xi_{c,m,n}^+ - \xi_{c,m,n}^- = \hat{\nu}_{1,c,m,n}, \ (c,m,n) \in S_1^{IN} ,
\end{equation}
where $\hat{\nu}_1$ is given by \eqref{eq:nu hat get from conv nu}. Note that, by \eqref{eq:dual var1}
\begin{equation}\label{eq xi and hat nu1}
    \begin{aligned}
        &\xi_{c,m,n}^+ \geq [\hat{\nu}_{1,c,m,n}]_+ \\
        &\xi_{c,m,n}^- \geq [\hat{\nu}_{1,c,m,n}]_- 
    \end{aligned} \ .
\end{equation}
As such, to maximize $-(\xi^+_{c,m,n}+\xi^-_{c,m,n})$ that appears in the objective function $g$ (cf. \eqref{eq:dual obj}), we choose $\xi_{c,m,n}^+ = [\hat{\nu}_{1,c,m,n}]_+$ and $\xi_{c,m,n}^- = [\hat{\nu}_{1,c,m,n}]_-$. Finally, we can simplify \eqref{dual} as \eqref{dual network}, which we refer as the CNN dual network.


{  
    %
\begin{subequations}\label{dual network}
    \begin{align}
        & \text{maximize}\nonumber \\
        \begin{split}
            \quad &g({\Theta})
            = \sum\limits_{(c,m,n) \in S_1^{IN}}-x_{c,m,n}\hat{\nu}_{1,c,m,n} 
            - \sum\limits_{(c,m,n) \in S_1^{IN}} \epsilon_{c,m,n}| \hat{\nu}_{1,c,m,n} | \\
            & \quad -\sum\limits_{l=1}^{k-1} \sum\limits_{(c,m,n)\in F_{l+1}^{IN}} \nu_{l+1,c,m,n}b_{l,c} + \sum\limits_{l=2}^{k}\sum\limits_{(c,m,n) \in \mathcal{I}_l} \hat{l}_{l,c,m,n}[\nu_{l,c,m,n}]_+ \\
            & \quad + \sum\limits_{l=2}^{k}\sum\limits_{(c,m,n,a)\in \bar{\mathcal{I}}_l} (\bar{l}_{l,c,m,n})_a [(\kappa_{l,c,m,n})_a]_+ -\sum\limits_{l=k}^{k+t-1} {\nu'_{l+1}}^T b_l + \sum\limits_{l=k+1}^{k+t-1}\sum\limits_{a \in \hat{\mathcal{I}}_l} l'_{l,a}[\nu'_{l,a}]_+ \\
        \end{split}\label{eq:dualNet obj} \\ 
        &\textbf{subject to}\nonumber \\
        \begin{split}
            \quad & \bm{\nu}'_{k+t} = \bm{d} \\ 
        \end{split}\label{eq:dualNet con1} \\
        \begin{split}
            \quad & \hat{\nu}'_{l,a} = (\bm{W}_l^T \bm{\nu}'_{l+1})_a , \ ~ { 
            \left\{ \begin{aligned}
                &l \in \llbracket k+1,k+t-1 \rrbracket, \\
                &a \in \llbracket 0,a_l-1 \rrbracket
            \end{aligned}
            \right .} \\
        \end{split}\label{eq:dualNet con2} \\
        \begin{split}
            \quad & \nu'_{l,a} = \left \{
                \begin{aligned}
                    & 0 &&, a \in \hat{\mathcal{I}}_l^- \\
                    & \hat{\nu}'_{l,a} &&, a \in \hat{\mathcal{I}}_l^+ \\ 
                    & \frac{u'_{l,a}}{u'_{l,a}-l'_{l,a}}[\hat{\nu}'_{l,a}]_+ - \alpha'_{l,a}[\hat{\nu}'_{l,a}]_- &&, a \in \hat{\mathcal{I}}_l \\
                \end{aligned}
            \right \}, { l \in \llbracket k+1,k+t-1 \rrbracket} \\ 
        \end{split}\label{eq:dualNet con3} \\
        \begin{split}
            \quad & \bm{\gamma}_k = \bm{W}_k^T \bm{\nu}'_{k+1} \\
        \end{split}\label{eq:dualNet con4} \\
        \begin{split}
            \quad & \left \{
                \begin{aligned}
                    & \beta_{k,c,m,n} = \sum\limits_{a=0}^{a_k-1} (\hat{W}_{c,m,n})_a \gamma_{k,a} &&, (c,m,n) \in S_k^{IN} \\
                    & \beta_{l,c,m,n} = \hat{\nu}_{l,c,m,n}  &&, \  {
                    \left\{ 
                    \begin{aligned}
                        &l \in \llbracket 2,k-1 \rrbracket, \\
                        &(c,m,n) \in S_l^{IN}
                    \end{aligned}
                    \right .}\\
                \end{aligned} 
            \right . \\
        \end{split}\label{eq:dualNet con5} \\
        \begin{split}
            \quad & (\rho_{l,c,m,n})_{k_l^M k_l^M} = \beta_{l,c,m,n}, \ 
            {
                    \left\{ 
                    \begin{aligned}
                        &l \in \llbracket 2,k \rrbracket, \\
                        &(c,m,n) \in S_l^{IN}
                    \end{aligned}
                    \right .}\\ 
        \end{split}\label{eq:dualNet con6} \\
        \begin{split}
            \quad & (\kappa_{l,c,m,n})_a = \left \{
                \begin{aligned}
                    & 0 &&, (c,m,n,a) \in \bar{\mathcal{I}}_l^- \\
                    & (\rho_{l,c,m,n})_{a+1} &&, (c,m,n,a) \in \bar{\mathcal{I}}_l^+ \\ 
                    & \frac{(\bar{u}_{l,c,m,n})_a}{(\bar{u}_{l,c,m,n})_a-(\bar{l}_{l,c,m,n})_a}[(\rho_{l,c,m,n})_{a+1}]_+ - (\alpha_{l,c,m,n}^M)_a[(\rho_{l,c,m,n})_{a+1}]_- &&, (c,m,n,a) \in \bar{\mathcal{I}}_l \\
                \end{aligned}
            \right \} \\
            &\quad \quad \quad \quad \quad , { l \in \llbracket 2,k \rrbracket }\\ 
        \end{split}\label{eq:dualNet con7} \\
        \begin{split}
            \quad &(\rho_{l,c,m,n})_a = (\rho_{l,c,m,n})_{a+1} - (\kappa_{l,c,m,n})_a,\ {
                    \left\{ 
                    \begin{aligned}
                        &l \in \llbracket 2,k \rrbracket, \\
                        &(c,m,n) \in S_l^{IN}, \\
                        & a \in \llbracket 1 , k_l^Mk_l^M-1 \rrbracket
                    \end{aligned}
                    \right .}\\
        \end{split}\label{eq:dualNet con8} \\
        \begin{split}
            \quad &\hat{\kappa}_{l,c,m,n} = \sum\limits_{\substack{(i,j) \in \llbracket 0, k_l^M-1 \rrbracket \times \llbracket 0, k_l^M-1 \rrbracket \\ (c,\frac{m-i}{s_l^M},\frac{n-j}{s_l^M}) \in S_l^{IN}}} (\kappa_{l,c,\frac{m-i}{s_l^M},\frac{n-j}{s_l^M}})_{i k_l^M+j} , \  {
                    \left\{ 
                    \begin{aligned}
                        &l \in \llbracket 2,k \rrbracket, \\
                        &(c,m,n) \in Q_l^{IN}
                    \end{aligned}
                    \right .}\\
        \end{split}\label{eq:dualNet con9} \\
        \begin{split}
        \quad & \nu_{l,c,m,n}= \left \{
                \begin{aligned}
                    & 0 &&, (c,m,n) \in \mathcal{I}_l^- \\
                    & \hat{\kappa}_{l,c,m,n} &&, (c,m,n) \in \mathcal{I}_l^+ \\ 
                    & \frac{\hat{u}_{l,c,m,n}}{\hat{u}_{l,c,m,n} - \hat{l}_{l,c,m,n}}[\hat{\kappa}_{l,c,m,n}]_+ - \alpha_{l,c,m,n}^R[\hat{\kappa}_{l,c,m,n}]_- &&, (c,m,n) \in \mathcal{I}_l \\
                \end{aligned}
            \right \}, \ { l \in \llbracket 2,k \rrbracket} \\
        \end{split}\label{eq:dualNet con10} \\
        \begin{split}
            \quad & \hat{\nu}_{l,c,m,n} = \sum\limits_{(\hat{c},m',n') \in F_{l+1}^{IN}} \nu_{l+1,\hat{c},m',n'} h_{l,\hat{c}}({c,m+p_l^{cv}-s_l^{cv}m',n+p_l^{cv}-s_l^{cv}n'}), \ { 
                            \left \{
                            \begin{aligned}
                                & l\in \llbracket 1,k-1 \rrbracket \\
                                & (c,m,n) \in S_l^{IN} \\
                            \end{aligned}
                        \right .} \\
        \end{split}\label{eq:dualNet con11} 
    \end{align}
\end{subequations}
}

\subsection{Bound analysis for intermediate layers}
\label{section:Bound Analysis for Intermediate Layers}

From Fig.~\ref{fig:compare cifar10}, we observe that the accuracy lower bound predicted by each verification method becomes looser as $\epsilon$ increases, and such phenomenon is also present in CAPM.
As bounds estimated for the previous operations will affect the later operation’s bound prediction, it is crucial for us to understand how the bounds estimated in each intermediate operation (e.g., convolution, ReLU, Maxpool) differ from the actual adversarial polytopes. As such we may identify the dominant factors that loosens the accuracy lower bound as a guidance for future improvements.


\subsubsection{Evaluate bound tightness by bound gap}\label{subsubsec:Evaluate Bound Tightness by Bound Gap}
If the real upper/lower bounds for each neuron {from} an arbitrary adversarial example were known, then a comparison with those real bounds would demonstrate the accuracy of the predicted bounds for each intermediate operation. The gap between the upper and lower bounds for the real and predicted bounds are compared, which is described as
\begin{gather*}
    g_{r}=u_{r}-l_{r} \quad \text{and} \quad
    g_{p}=u_{p}-l_{p}
\end{gather*}
where $u_{r}$ and $l_{r}$ denote the real upper and lower bounds, and $u_{p}$ and $l_{p}$ denote the predicted upper and lower bounds. The difference in the bound gap $g_{p}-g_{r}$ is always non-negative and a perfectly predicted bound gives a zero bound gap difference. 
By reporting the real/predicted bound gaps averaged over all neurons pertaining to that operation, we can demonstrate how tight the predicted bounds are in general compared to the real bounds in each specific operation.


In reality, the real upper and lower bounds are not known. A Monte-Carlo simulation procedure for an adversary randomly draws a large amount of $N_{adv}$ adversary examples from the adversarial polytope at the input layer following a uniform distribution. The $N_{adv}$ randomly drawn adversarial examples are then fed into the network and the intermediate values for each neuron are computed. 
For each neuron, the maximum and minimum of the $N_{adv}$ values as realized by the $N_{adv}$ adversary examples are then computed as an estimate of the real upper and lower bounds.

\begin{figure*}
    \centering
    \includegraphics[width=0.7\textwidth]{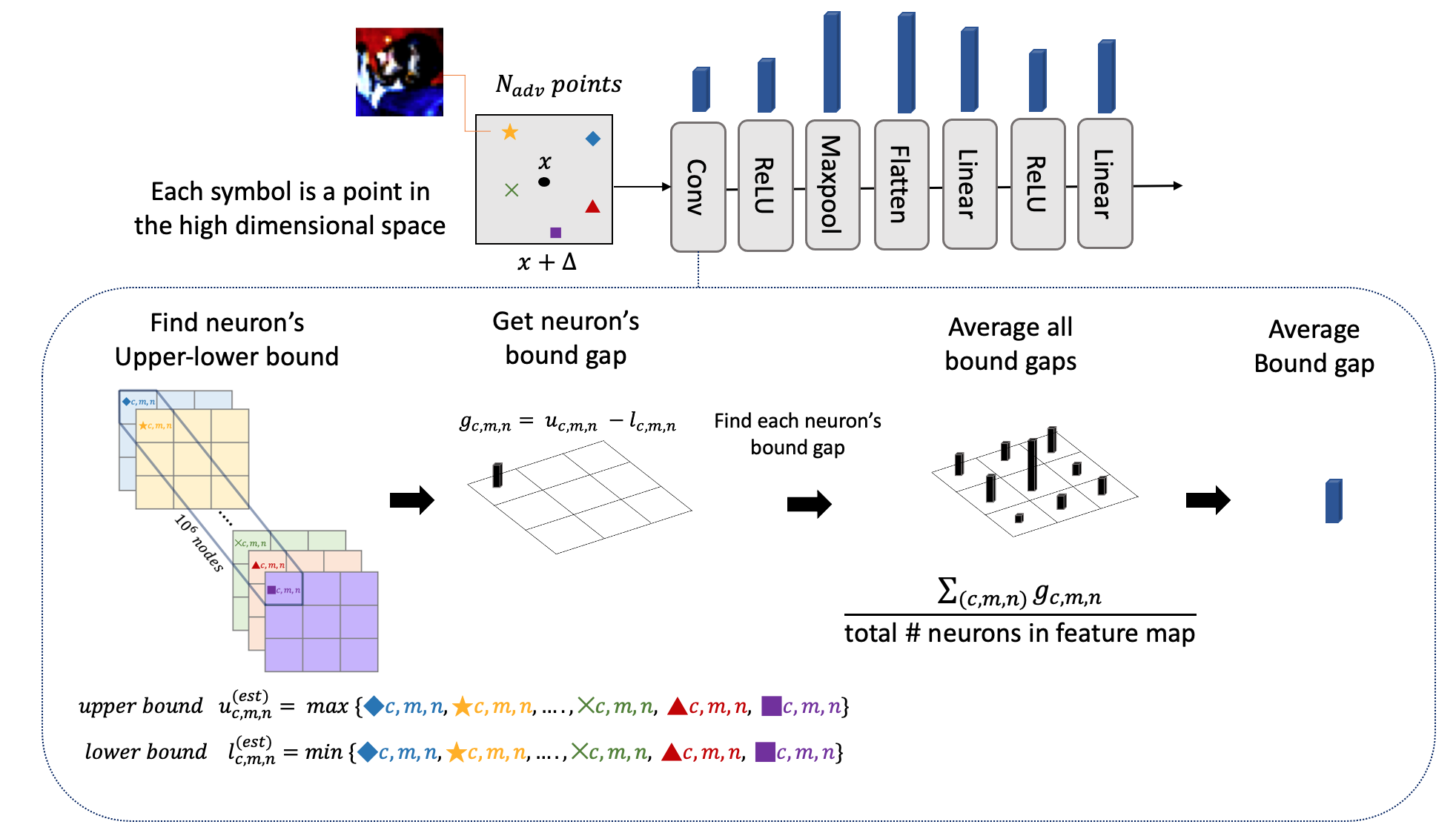}
    \caption{
    Estimating the real upper and lower bounds using a Monte Carlo simulation for an adversarial polytope.
    %
    } 
    \label{fig:simulate polytope process}
\end{figure*}

Fig.~\ref{fig:simulate polytope process} illustrates how the real upper and lower bounds are estimated through the simulated adversarial polytope.
The gray square at the upper left corner is the adversarial polytope $\{x + \Delta: \lVert \Delta \rVert_\infty \leq \epsilon\}$ in some high dimensional input space, where the black dot represents the original image $x$. We draw $N_{adv}$ points from the adversarial polytope, as indicated by the colorful symbols in the gray square, which are then fed into the network and resulted in $N_{adv}$ feature maps after the convolution operation. For a specific neuron that corresponds to the $(m,n)$'th pixel in the $c$'th channel of the feature map, this amounts to $N_{adv}$ values, to which the maximum $u^{(est)}_{c,m,n}$ and minimum $l^{(est)}_{c,m,n}$ are computed as an approximation to the real upper and lower bounds. We can then estimate the real bound gap as $g^{(est)}_{c,m,n}=u^{(est)}_{c,m,n}-l^{(est)}_{c,m,n}$ which is then compared with the predicted bound gap. \yw{However, since the Monte-Carlo Simulation do not reach a tight approximation, we leave the further discussion in appendix \ref{subsubsec:Verify the idea of simulating polytope}.}

\subsubsection{Limitation on approximating adversarial polytope with Monte Carlo simulation}\label{subsubsec:Verify the idea of simulating polytope}

In the previous section we described an intuitive and easy-to-implement Monte-Carlo simulation method to estimate the real upper and lower bounds of the intermediate adversarial polytopes at each operation.  Several questions naturally arise: Under what circumstances does such Monte-Carlo simulation method yield a good estimate to the real upper and lower bounds to the adversarial polytopes? Is it possible that such Monte-Carlo simulation method can even replace the optimization-theoretic robustness verification methods proposed in literature as well as in this work?
To answer such questions, we consider a simple neural network consisting of only one fully-connected layer, to which the lower bound of the $j$-th output node subject to adversarial examples with $l_\infty$ norm-bounded constraint is described as the following optimization problem: (The upper bound can be formulated in an analogous manner, namely by replacing $e_j$ with $-e_j$ followed by a negation.)
\begin{equation}\label{extend lower bound problem}
    \begin{aligned}
        \min_{\bf{y} } \quad & e_j^T y\\
        \textrm{s.t.} \quad & z_1 \leq x+\epsilon \\
                      \quad & z_1 \geq x-\epsilon \\
                      \quad & y = W_1^T z_1 + b_1 \\ 
    \end{aligned}
\end{equation}
Since (\ref{extend lower bound problem}) is a convex optimization problem, its primal and dual optimums must coincide. {We use the CVX tool \citep{cvxpy} to find the primal optimum, and apply the dual network (cf. \citep{wong2018provable}, also in Sec.~\ref{subsection:Overview of our method}) to find dual optimum.} The primal/dual optima are then compared with the bounds that re-estimated using the Monte-Carlo simulation method in Sec.~\ref{subsubsec:Evaluate Bound Tightness by Bound Gap}.

\begin{figure}[ht]
\centering
\includegraphics[width=0.45\textwidth]{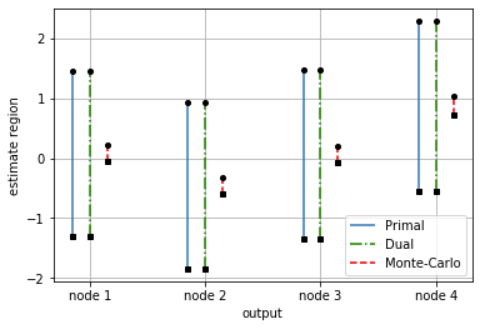}
\caption{\label{fig:much_nodes_input}
Comparison of (estimated) upper/lower bounds for the output nodes using the primal approach, the dual approach and a Monte-Carlo simulation with a one-layer network with 784-dimensional inputs. The endpoints of each line represent the upper (circle) and lower (square) bounds that are respectively predicted by each method.
%
}
\end{figure}

\begin{figure}[ht]
\centering
\includegraphics[width=0.45\textwidth]{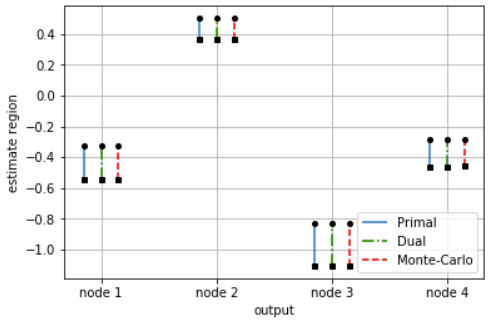}
\caption{\label{fig:less nodes input}
Comparison of (estimated) upper/lower bounds for the output nodes using the primal approach, the dual approach and a Monte-Carlo simulation with a one-layer network with 3-dimensional inputs. The endpoints of each line represent the upper (circle) and lower (square) bounds that are respectively predicted using each method.
%
}
\end{figure}

Fig.~\ref{fig:much_nodes_input} shows a network consisting of one fully-connected linear layer with 784-dimensional inputs and 4-dimension outputs. $N_{adv} = 10^6$ points are randomly sampled from the 784-dimensional cube $[0,1)^{784}$, and set $\epsilon = 0.1$, which corresponds to a large adversarial polytope, using the experimental settings in DeepPoly \citep{singh2019abstract}. Fig.~\ref{fig:much_nodes_input} shows that both the primal and dual optima coincide, so there is a perfect zero duality gap, as expected for a convex optimization problem.
However, there is a significant gap between the primal/dual optima and the bound that is estimated using a Monte-Carlo simulation. This demonstrates that even $10^6$ Monte-Carlo simulations are not sufficient to properly represent the complicated details of an actual adversarial polytope that may affect the bounds. This is intuitively reasonable because sampling the corners of the input adversarial polytope (which is a $784$-dimensional
cube) requires $2^{784} \approx 10^{236}$ samples which is not practically possible. If the input dimension is small (3 dimensions), then the bound that is estimated using a Monte-Carlo simulation coincides with the primal and dual optima, as shown in Fig.~\ref{fig:less nodes input}.
%
A Monte-Carlo simulation method may provide us some insight on how well the bound predicted by a verification method is (cf. Sec.~\ref{subsubsec:Evaluate Bound Tightness by Bound Gap}),  but  Monte-Carlo simulation does not accurately represent the complicated details of the adversarial polytope, especially in highly dimensional input settings, so the bound gap for the actual adversarial polytope is underestimated (cf. Sec.~\ref{subsubsec:Evaluate Bound Tightness by Bound Gap}). This further highlights the necessity of a provable robustness verification scheme as discussed in this work.
\subsection{Network structure} \label{appendices: network structure}
For the CNNs verified in Sec.~\ref{section:experiment} (namely convSmallMNIST, convMedMNIST, convBigMNIST, convSmallCIFAR10, convMedCIFAR10, and convBigCIFAR10), we add maxpool layers to the convSmall, convMed, and convBig counterparts in \citep{mirman2018differentiable} and slightly adjust the parameters of striding and padding to achieve similar number of parameters as in \citep{mirman2018differentiable}. The detailed network architectures are listed below:
\noindent
\textbf{convSmallMNIST} \\
$input$ $\rightarrow$ $Conv_{2,1,2, 1} \ 16 \times 4 \times 4$ $\rightarrow$ $Conv_{2,1,2, 1} \ 32 \times 4 \times 4$ $\rightarrow$ $Flat$ $\rightarrow$ $FC(800,100)$ $\rightarrow$ $ReLU$ $\rightarrow$ $FC(100,10)$ $\rightarrow$ $output$ \\
\\\textbf{convMedMNIST} \\
$input$ $\rightarrow$ $Conv_{1,1,2, 0} \ 16 \times 2 \times 2$ $\rightarrow$ $Conv_{1,1,2, 0} \ 32 \times 2 \times 2$ $\rightarrow$ $Flat$ $\rightarrow$ $FC(1568,100)$ $\rightarrow$ $ReLU$ $\rightarrow$ $FC(100,10)$ $\rightarrow$ $output$ \\
\\\textbf{convBigMNIST} \\
$input$ $\rightarrow$ $Conv_{1,1,2, 1} \ 32 \times 3 \times 3$ $\rightarrow$ $Conv_{1,0,2, 1} \ 32 \times 4 \times 4$ $\rightarrow$ $Conv_{1,0,2, 1} \ 64 \times 3 \times 3$ $\rightarrow$ $Conv_{2,0,2, 0} \ 64 \times 4 \times 4$ $\rightarrow$ $Flat$ $\rightarrow$ $FC(1024,512)$ $\rightarrow$ $ReLU$ $\rightarrow$ $FC(512,512)$ $\rightarrow$ $ReLU$ $\rightarrow$ $FC(512,10)$ $\rightarrow$ $output$ \\
\\\textbf{conv}$_{\mathbf{S}}$ \textbf{MNIST} \\
$input$ $\rightarrow$ $Conv_{2,1,2, 2} \ 16 \times 4 \times 4$ $\rightarrow$ $Conv_{2,1,2, 2} \ 32 \times 4 \times 4$ $\rightarrow$ $Flat$ $\rightarrow$ $FC(32,24)$ $\rightarrow$ $ReLU$ $\rightarrow$ $FC(24,10)$ $\rightarrow$ $output$ \\
\\\textbf{conv}$_{\mathbf{M}}$ \textbf{MNIST} \\
$input$ $\rightarrow$ $Conv_{2,1,2, 2} \ 16 \times 4 \times 4$ $\rightarrow$ $Conv_{2,1,2, 2} \ 32 \times 4 \times 4$ $\rightarrow$ $Conv_{1,1,2, 2} \ 64 \times 2 \times 2$ $\rightarrow$ $Flat$ $\rightarrow$ $FC(64,32)$ $\rightarrow$ $ReLU$ $\rightarrow$ $FC(32,10)$ $\rightarrow$ $output$ \\
\\\textbf{conv}$_{\mathbf{L}}$ \textbf{MNIST} \\
$input$ $\rightarrow$ $Conv_{1,1,2, 2} \ 32 \times 2 \times 2$ $\rightarrow$ $Conv_{1,1,2, 2} \ 64 \times 2 \times 2$ $\rightarrow$ $Conv_{1,1,2, 2} \ 128 \times 2 \times 2$ $\rightarrow$ $Flat$ $\rightarrow$ $FC(2048,256)$ $\rightarrow$ $ReLU$ $\rightarrow$ $FC(256,10)$ $\rightarrow$ $output$ \\
\\\textbf{convSmallCIFAR10} \\
$input$ $\rightarrow$ $Conv_{2,1,2, 1} \ 16 \times 4 \times 4$ $\rightarrow$ $Conv_{2,1,2, 1} \ 32 \times 4 \times 4$ $\rightarrow$ $Flat$ $\rightarrow$ $FC(1152,100)$ $\rightarrow$ $ReLU$ $\rightarrow$ $FC(100,10)$ $\rightarrow$ $output$ \\
\\\textbf{convMedCIFAR10} \\
$input$ $\rightarrow$ $Conv_{1,1,2, 0} \ 16 \times 2 \times 2$ $\rightarrow$ $Conv_{1,1,2, 0} \ 32 \times 2 \times 2$ $\rightarrow$ $Flat$ $\rightarrow$ $FC(2048,100)$ $\rightarrow$ $ReLU$ $\rightarrow$ $FC(100,10)$ $\rightarrow$ $output$ \\
\\\textbf{convBigCIFAR10} \\
$input$ $\rightarrow$ $Conv_{1,1,2, 1} \ 32 \times 3 \times 3$ $\rightarrow$ $Conv_{1,1,2, 0} \ 32 \times 4 \times 4$ $\rightarrow$ $Conv_{1,0,2, 1} \ 64 \times 3 \times 3$ $\rightarrow$ $Conv_{2,0,2, 1} \ 64 \times 4 \times 4$ $\rightarrow$ $Flat$ $\rightarrow$ $FC(4096,512)$ $\rightarrow$ $ReLU$ $\rightarrow$ $FC(512,512)$ $\rightarrow$ $ReLU$ $\rightarrow$ $FC(512,10)$ $\rightarrow$ $output$ \\
Here $Conv_{s,p,k,t} \ C \times W \times H$ represents a convolution layer consisting of a convolution operation followed by $ReLU$ and maxpool with kernel size $k \times k$ and stride $t$, where the convolution operation consists of $C$ convolution kernels each of width $W$ and height $H$ along with stride $s$ and padding $p$; $FC(M,N)$ represents a fully-connected layer with $M$ input neurons and $N$ output neurons
We use open source repo provided in \citep{wong2020fast} for adversarial training and slightly modify its default parameters to fit our models. Specifically, the hyperparameters we used are illustrated in Table \ref{tab: hyperparameters}.\\
\begin{table*}[!b]
\caption{Hyperparameter for network training} \label{tab: hyperparameters}
\centering
\begin{tabular}{ |c|c|c|c|c|c|c|c| }
\hline
    Dataset         & Method & epochs & batch size & learning rate & Optimizer  & $\epsilon$ & alpha\\
 \hline 
             ~  & Normal & 200 & 500 & 0.0001 & Adam & - & - \\
             MNIST  & Fast & 200 & 500 & max 0.005 & SGD & 8/255 & 2/255 \\
             ~  & PGD & 200 & 500 & max 0.005 & SGD & 8/255 & 2/255 \\
 \hline
             ~  & Normal & 200 & 500 & 0.0001 & Adam & - & - \\
             CIFAR10  & Fast & 200 & 500 & max 0.005 & SGD & 8/255 & 2/255 \\
             ~  & PGD & 200 & 500 & max 0.005 & SGD & 2/255 & 2/255 \\
 \hline
\end{tabular}
\end{table*}
\subsection{Reproducing the state-of-the-art method}\label{sec:reproduce_sota}
\begin{figure}[h]
\centering
\includegraphics[width=0.8\textwidth]{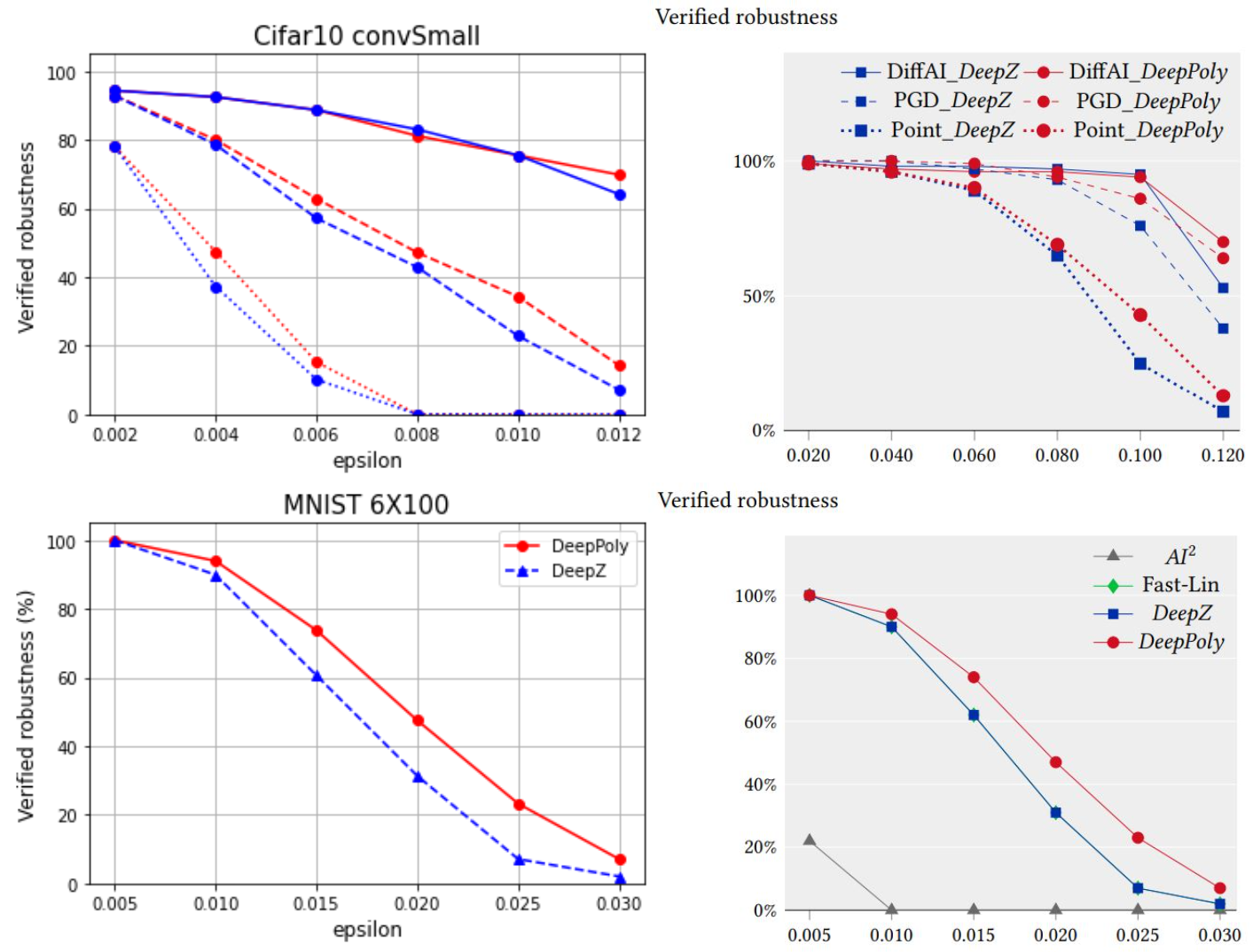}
\label{fig:reproduce_result}
\caption{Verification results for DeepPoly and DeepZ using the settings in \citep{singh2019abstract,singh2018fast} for the FFNNSmall and ConvSmall models \citep{singh2019abstract}. The upper left figure shows the reproduced result for ConvSmall for Cifar10, which is consistent to the reported result (upper right) in the Fig.11(a) of \citep{singh2019abstract}. The bottom left figure shows the reproduced result for FFNNSmall for MNIST $6X100$, which is consistent to the reported result (bottom right) in the Fig.5(a) of \citep{singh2019abstract}.}
\end{figure}

As there were no reported empirical verification performance on maxpool-based CNNs, it is necessary to correctly reproduce and test the state-of-the-art methods on maxpool-based CNNs such as the models described in Sec.~\ref{subsec:Verification of CNN with maxpooling}.
We reproduce DeepPoly and DeepZ on the neural networks (which includes convolution and ReLU but without maxpool) following the settings in \citep{singh2019abstract,singh2018fast} with the implementation provided in \citep{eth-sri}. Fig.~\ref{fig:reproduce_result} indicates that the reproduced result is highly consistent with the reported results in \citep{singh2019abstract}. 

PRIMA is reproduced using a pretrained model from \citep{eth-sri}. Table \ref{tab: PRIMA reproduce} shows the reproduced results for verified robustness for the first $1000$ samples and the average verification time for convSmall and convBig. The difference in runtime is attributed to the use of different hardware but the verified robustness is highly consistent with the reported results in \citep{muller2022prima}.
%

\begin{table*}
\centering
\caption{Reproduced result for PRIMA for convSmall and convBig.}
    \label{tab: PRIMA reproduce}
\begin{tabular}{ |c|c|c|c|c|c|c|c|c| }
\hline
    Dataset        &   Model        & Training & Accuracy & $\epsilon$ & Ver & Time & \makecell[c]{Ver\\ (Reported)} & \makecell[c]{Time \\(Reported)} \\
 \hline 
 \hline
    MNIST & convSmall & NOR & 980 & 0.120 & 650 & 132 &640 & 51 \\
        ~ & convBig & DiffAI & 929 & 0.300 & 775 & 35.5 & 775 & 10.9 \\
 \hline
    CIFAR10 & convSmall & PGD & 630 & 2/255 & 459 & 24.4 & 458 & 16 \\
 \hline
\end{tabular}
\end{table*}

\end{document}